\documentclass[11pt]{article}
\usepackage[preprint]{acl}
\usepackage{times}
\usepackage{latexsym}
\usepackage[T1]{fontenc}
\usepackage[utf8]{inputenc}
\usepackage{inconsolata}
\usepackage{amsmath,amssymb}
\usepackage{booktabs}
\usepackage{graphicx}
\usepackage{capt-of}
\usepackage{microtype}
\usepackage{array}
\usepackage{xcolor}
\usepackage{multirow,colortbl}
\usepackage{hyperref}
\usepackage{xurl}
\usepackage{needspace}
\usepackage{placeins}
\usepackage[breakable,skins]{tcolorbox}
\definecolor{visepromptback}{RGB}{247,249,248}
\definecolor{viseauthorframe}{RGB}{88,142,126}
\definecolor{viseknowledgeframe}{RGB}{83,124,165}
\definecolor{visefeedbackframe}{RGB}{139,105,157}
\definecolor{viserevisionframe}{RGB}{145,110,70}
\definecolor{viseablationframe}{RGB}{150,90,90}
\tcbset{visepromptlayout/.style={breakable}}
\newtcolorbox{visepromptbox}[2][viseauthorframe]{
  enhanced jigsaw, breakable,
  colback=visepromptback, colframe=#1, colbacktitle=#1, coltitle=white,
  fonttitle=\bfseries\footnotesize, title={#2},
  boxrule=0.8pt, arc=2mm,
  left=5pt, right=5pt, top=4pt, bottom=4pt,
  before skip=5pt, after skip=5pt, pad at break*=1pt,
  fontupper=\footnotesize,
  before upper={\raggedright\setlength{\parindent}{0pt}},
  visepromptlayout
}

\newcommand{\system}{\textsc{VISE}}
\newcommand{\skillmd}{\texttt{SKILL.md}}
\title{VISE: Verifier-Independent Skill Evolution with Feedback\\from Tool Implementations}
\author{
\textbf{Yuxuan Liu$^{1}$ \quad Chongyang Du$^{2}$ \quad
Yuhao Zhang$^{2}$ \quad Jiahe Guo$^{3}$} \\
\textbf{Kawai Chung$^{1}$ \quad Wenbin Hu$^{1}$ \quad
Huihao Jing$^{1}$ \quad Hongyu Luo$^{1}$} \\
\textbf{Qing Zong$^{1}$ \quad Haochen Shi$^{1}$ \quad
Haoran Li$^{1\dagger}$ \quad Yangqiu Song$^{1}$} \\[3pt]
$^{1}$The Hong Kong University of Science and Technology, Hong Kong SAR, China \\
$^{2}$Harbin Institute of Technology, Shenzhen, China \\
$^{3}$Harbin Institute of Technology, Harbin, China \\
\texttt{\{yliurk,hlibt\}@connect.ust.hk, yqsong@cse.ust.hk} \\
$^{\dagger}$Corresponding author.
}
\begin{document}
\maketitle
\begin{abstract}
Agent skills provide reusable workflows that guide LLM agents in using
tools. Refining these workflows requires feedback on missing or incorrect
instructions, but task-specific verifiers are not always available.
We propose \system{} (Verifier-Independent Skill Evolution), a
scripts-first approach that uses tool implementations to guide skill
revision. It extracts tool capabilities, input requirements, usage
conditions, and output formats from source code. Using the task request
and execution trace, it compares this information with current workflow
instructions to identify missing or conflicting steps. It then recommends
specific edits to \skillmd{}, linking each change to the relevant
implementation and usage conditions. We evaluate \system{} as an
additional revision stage for four evolution methods on SkillsBench and
SkillLearnBench. Compared with the corresponding No-verifier conditions,
it improves overall average task success by 9.14--13.35 percentage points.
The proportion of tasks solved in all three runs increases by
12.36--18.29 points. A case study shows how revised instructions connect
tool outputs to later steps. These results support tool implementations
as a practical feedback source for skill self-evolution.
\end{abstract}
\section{Introduction}
\label{sec:intro}

Agent skills provide reusable procedures that help LLM agents use tools
to complete tasks~\citep{zhang2025agentskills,li2026skillsbench}.
Effective skills specify which tools to use, how to prepare their inputs,
and how to use their outputs. For example, an object-counting workflow
must explain which video frames to process and how to combine the
returned counts. Without this guidance, an agent may call a tool without
using its output to complete the task. Skill self-evolution refines such
instructions through execution experience and
feedback~\citep{zhang2026skillevolver,liu2026skillrevise}. Feedback must therefore
specify which instructions to revise and how.\looseness=-1

\begin{figure}[t]
  \centering
  \includegraphics[width=0.92\columnwidth]{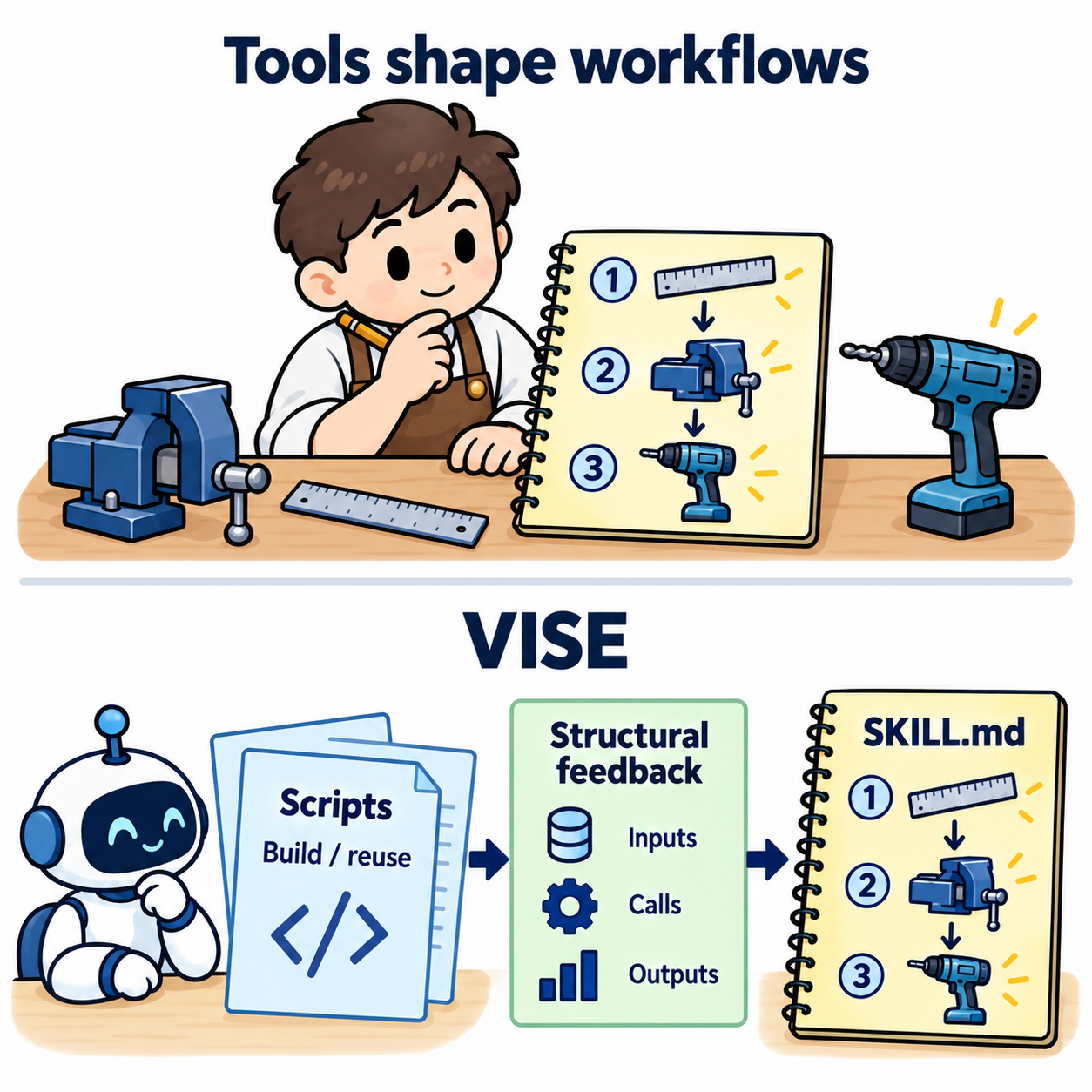}
  \caption{Scripts-first intuition behind VISE. People organize workflows
  around tools. VISE similarly builds or reuses scripts and derives structural
  feedback from their implementations to refine \texttt{SKILL.md}.}
  \label{fig:scripts_first_intuition}
\end{figure}

Existing approaches draw on model-generated feedback, execution
experience, and task evaluation. Self-Refine uses self-feedback to refine
outputs~\citep{madaan2023selfrefine}, while SkillLearnBench studies
self-feedback for skills~\citep{zhong2026skilllearnbench}. Reflexion
turns execution experience into verbal reflections that guide later
attempts~\citep{shinn2023reflexion}. Verifier-based methods use success
judgments or diagnostic reports to guide skill
updates~\citep{zhang2026skillevolver}. SkillRevise also re-executes
revised skills to select the best observed
candidate~\citep{liu2026skillrevise}. Related work extracts procedural
knowledge from code: Repo-To-Skill distills repository knowledge into
reusable skills~\citep{chen2026repotoskill}.

\newpage
When task-verifier feedback is unavailable, agents must determine how to
revise skills from task requirements, tool implementations, and execution
history. However, an available tool may support only part of a task or
require inputs and conditions not covered by existing instructions.
Effective revision therefore requires determining which tool
capabilities apply to the task and how their requirements should shape
the workflow. Agents must then identify where existing instructions omit
necessary steps or conflict with the implementation, and translate those
gaps into actionable revisions.

We propose \system{} (Verifier-Independent Skill Evolution), a
scripts-first approach that uses tool implementations to guide skill
revision. For new skills, it first selects or creates task-relevant
scripts and then writes initial instructions in \skillmd{}. For revision,
it extracts information about tool capabilities, usage conditions,
steps, and constraints, with references to supporting source code.
It compares this information with the current instructions using the
task request and execution trace. For each missing or conflicting step,
it recommends a specific change and states when it applies. The revision
stage updates the relevant passages in \skillmd{} while retaining the
scripts. This stage can also refine skills produced by other evolution
methods. Figure~\ref{fig:scripts_first_intuition} illustrates the approach.

We evaluate \system{} on SkillsBench and
SkillLearnBench~\citep{li2026skillsbench,zhong2026skilllearnbench} by
further revising the No-verifier outputs of Self-Refine, TraceEvo,
SkillLearn, and SkillRevise. Across the four methods, overall average
task success increases by \mbox{9.14--13.35\%} over No-verifier.
The proportion of tasks solved in all three runs improves by
\mbox{12.36--18.29\%} (Table~\ref{tab:main}). \system{} also exceeds the
Binary, Report, and Diagnostic conditions on both overall metrics.
A video-counting case study illustrates how revised instructions connect
tool outputs to later steps. In a separate study, repeated evaluations
show that similar overall gains can hide changes in which tasks benefit.
Newly constructed SkillsBench variants further show that changing
requirements can reduce or reverse the skill benefits observed on the
original tasks.

Our contributions are threefold:
\par\noindent\textbf{1. Tool-grounded skill evolution.}
We propose \system{}, which links information from tool implementations
to missing or conflicting workflow instructions and generates targeted
revision feedback.
\par\noindent\textbf{2. A systematic comparison of feedback sources.}
We compare feedback sources across four evolution methods on SkillsBench
and SkillLearnBench, measuring average task success, task coverage, and
repeated execution success.
\par\noindent\textbf{3. An empirical study of skill utility.}
Through repeated evaluations and task variants, we analyze which tasks
benefit from skill use and how these benefits change with task
requirements.

\Needspace{7\baselineskip}
\section{Related Work}

\paragraph{Skill Self-Evolution.}
SkillLearnBench~\citep{zhong2026skilllearnbench} studies how agents revise
skills through self-feedback without external supervision.
SkillEvolver~\citep{zhang2026skillevolver} revises
instructions and code after deployment failures, while SPARK~\citep{zhou2026spark}
distills skills from environment-verified trajectories.
SkillRevise~\citep{liu2026skillrevise} combines defect diagnosis with retrieved
repair principles and re-executes revised skills to select the candidate with
the highest observed task utility.
Task evaluation also guides candidate selection in EvoSkill~\citep{alzubi2026evoskill}
and gates bounded skill edits in SkillOpt~\citep{yang2026skillopt}.
CoEvoSkills~\citep{zhang2026coevoskills} combines surrogate verification with
oracle feedback during evolution.
Multi-round analyses show that skill
improvements are sparse and depend on the feedback shown during
revision~\citep{liu2026rethinking}. These approaches
center on execution-derived feedback rather than tool implementations for revising
existing workflow guidance.

\paragraph{Tool-Grounded Procedural Knowledge.}
LATM~\citep{cai2024latm} and
CREATOR~\citep{qian2023creator} separate tool creation from use.
Voyager~\citep{wang2023voyager} builds an executable skill library,
while AutoManual~\citep{chen2024automanual} organizes
interaction-derived rules into planning manuals. DRAFT~\citep{qu2025draft} and
Play2Prompt~\citep{fang2025play2prompt} refine tool documentation through exploratory
interactions, with Play2Prompt generating tool-use examples for validation.
DocsChisel~\citep{lu2026docschisel} revises documentation fields using failed
execution traces and selects revisions by validation performance.
Repo-To-Skill~\citep{chen2026repotoskill} distills repository knowledge into skills
specifying capabilities, usage conditions, and procedures.
SkillScriptBench~\citep{liu2026skillscriptbench} evaluates documentation repair,
script repair, and preservation separately. Together, these studies examine
procedural knowledge through acquisition, documentation refinement,
package maintenance, and workflow evaluation.

\Needspace{7\baselineskip}
\section{An Empirical Study of Skill Utility}
\label{sec:motivation}

\begin{table}[t]
\centering\small
\begin{tabular*}{\columnwidth}{@{\extracolsep{\fill}}lrrr@{}}
\toprule
Method & Original & $|\Delta|$ & Variants \\
\midrule
No-skill & 41.70 & 8.08 & 41.09 \\
Skill-Creator & 37.70 & 8.91 & 41.76 \\
\midrule
SkillEvolver & 62.78 & 0.87 & 53.81 \\
SPARK & 38.78 & 4.30 & 42.26 \\
CoEvoSkills & 49.44 & 6.00 & 36.60 \\
Self-Refine & 60.46 & 0.16 & 58.55 \\
SkillLearn & 55.11 & 1.36 & 57.80 \\
Trace Evo & 61.82 & 3.87 & 58.84 \\
SkillRevise & 57.57 & 0.28 & 54.36 \\
\bottomrule
\end{tabular*}
\caption{Mean rewards (0--100) on SkillsBench and its variants.
Original averages two evaluations; $|\Delta|$ is their absolute difference
(Appendix~\ref{app:diagnostic}).}
\label{tab:motivation_combined}
\end{table}

Task-specific verifiers may be unavailable when skills need to be revised,
motivating the study of alternative feedback sources. Assessing these approaches
requires distinguishing execution performance from the benefit of using a skill.
We examine this distinction through repeated evaluations on SkillsBench~\citep{li2026skillsbench} and task variants
that modify input representations, task instances, or coupled
requirements.

\FloatBarrier

\subsection{Sources of Performance Gains}

A lower execution score does not necessarily imply a smaller benefit from using a skill.
CoEvoSkills scores decrease from 52.44 to 46.44 across two evaluations, while its
gains over the paired No-skill references remain similar, at 10.24 and 9.96 points.
Similar aggregate results can also conceal changes in which tasks succeed.
Self-Refine receives full reward on 50 tasks in each evaluation, yet nine tasks
lose full reward and nine others gain it. Additional task-level comparisons
appear in Appendix~\ref{app:gain_recurrence}.

\subsection{Utility under Changing Task Demands}
\label{sec:task_variants}

\begin{figure}[t]
\centering
\includegraphics[width=\columnwidth]{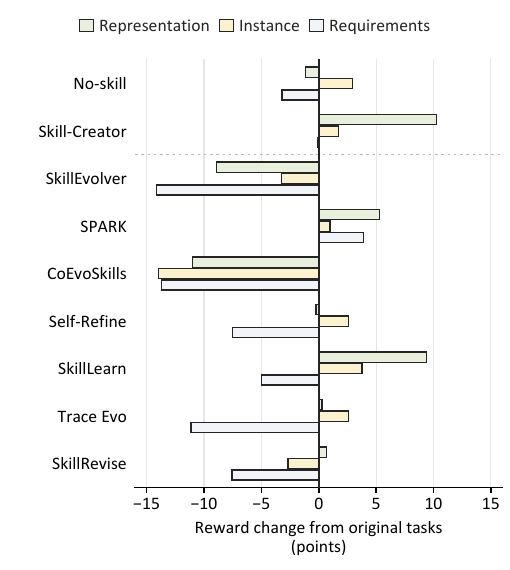}
\caption{Reward changes across task-change categories. Each bar subtracts the
method's Original mean in Table~\ref{tab:motivation_combined} from its
variant-category mean.}
\label{fig:motivation_analysis}
\end{figure}

We construct three types of SkillsBench variants to test whether skills remain
effective beyond their original tasks.
\emph{Representation changes (P1)} modify input or output representations, such as
file paths, field names, or serialization formats. \emph{Instance changes (P2)}
replace concrete task instances, such as data records or authorization
relationships. \emph{Coupled-requirement changes (P3)} alter interdependent
constraints or outputs that must remain consistent.
Each method's skills are generated or evolved on the original tasks and
evaluated on the variants without further revision.

As shown in Figure~\ref{fig:motivation_analysis},
SkillEvolver and CoEvoSkills score below their original-task averages in all three
categories, whereas SPARK scores above them. For Self-Refine, SkillLearn, Trace Evo,
and SkillRevise, coupled-requirement variants show the largest declines.
Nevertheless, these four methods retain higher overall variant scores than both
Skill-Creator and No-skill, \mbox{as shown in Table~\ref{tab:motivation_combined}}.

\begin{figure*}[t]
  \centering
  \includegraphics[width=\textwidth]{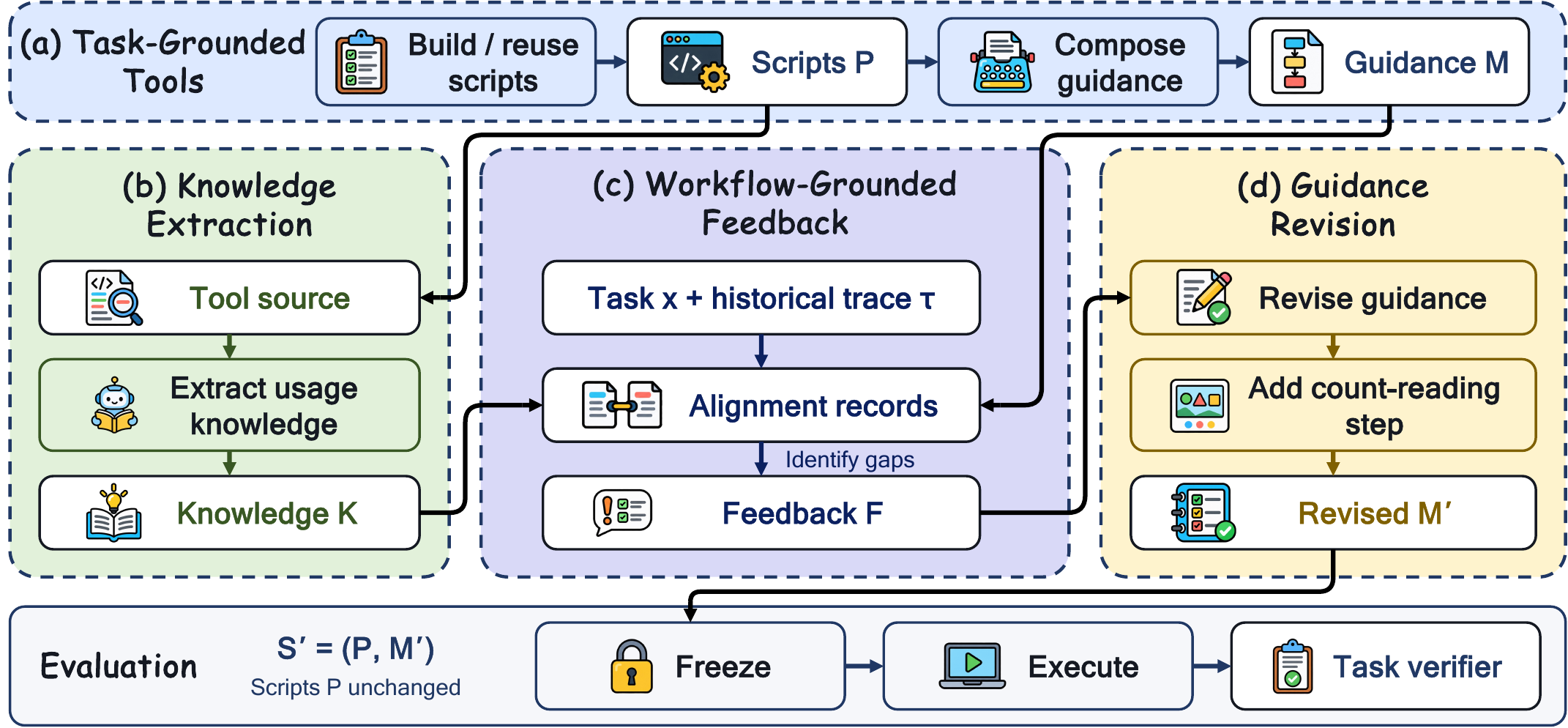}
  \caption{VISE uses executable tools to construct and revise workflow guidance.
  Scripts $P$ support initial guidance $M_0$; revision can start from $M_0$ or an
  evolved version $M$. Operational knowledge $\mathcal{K}$ is aligned with $M$
  under task $x$ and trace $\tau$ to identify guidance gaps and form feedback
  $\mathcal{F}$ for targeted edits, illustrated by adding per-pair count handling.
  Revised guidance $M'$ and unchanged scripts $P$ form $S'$, frozen before
  downstream evaluation.}
  \label{fig:scripts_first_pipeline}
\end{figure*}

\subsection{Revision Evidence and Research Questions}
These findings motivate assessing skill revision through both aggregate performance
and task-level benefits. Constructing a revision additionally requires identifying
what should change in the workflow: which operations to invoke, how to prepare
their inputs, and how to use their outputs. Public tool implementations expose
these operations and their constraints, providing a basis for examining the
corresponding workflow guidance. We investigate whether this implementation
knowledge can support effective skill revision without task-verifier feedback
through the following research questions.

\begingroup
\makeatletter
\renewcommand{\paragraph}{\@startsection{paragraph}{4}{\z@}%
  {.8ex \@plus .2ex}{-1em}{\normalfont\normalsize\bfseries}}
\makeatother
\paragraph{RQ1.}
How do feedback sources compare across evolution methods and task categories?

\paragraph{RQ2.}
Can tool-grounded feedback improve skills without task-verifier feedback?

\paragraph{RQ3.}
How is implementation evidence reflected in revised guidance and tool use?
\par\endgroup

\section{VISE: Evolving Workflows from Tools}
\label{sec:method}

\system{} follows a scripts-first paradigm for verifier-independent skill
evolution: executable tools provide the operations around which workflows are
built, while their implementations provide evidence for revising workflow
guidance. We represent a skill as $S=(P,M)$, comprising executable scripts $P$
and guidance $M$ in \skillmd{}. \system{} first establishes $P$ and composes
initial guidance around its operations. It then extracts operational knowledge
from $P$, compares this knowledge with $M$ in the task context, and constructs
structured feedback for targeted revision without task-verifier judgments.
Figure~\ref{fig:scripts_first_pipeline} illustrates the four stages.\looseness=-1

\subsection{Constructing Task-Grounded Tools}
\system{} constructs an initial skill by first establishing the executable
operations around which its workflow will be organized. Given a task request
$x$, it builds a script set $P$ by reusing suitable implementations or generating
missing scripts. These scripts implement concrete task operations and expose
their inputs, configurable parameters, and output formats, providing the
operational basis for composing workflow guidance.

{\widowpenalty=10000
\system{} then generates initial guidance $M_0$ from $x$ and $P$. The guidance
specifies how to prepare inputs for the available operations, arrange their
invocation, and use intermediate outputs in subsequent steps. Scripts implement
the operations, while the guidance coordinates their use to address the task
request. Together, they form the initial skill $S_0=(P,M_0)$.\par}

\subsection{Extracting Tool-Grounded Operational Knowledge}
Initial guidance $M_0$ organizes executable operations into a task workflow.
To support subsequent revision, \system{} extracts operational knowledge from
scripts $P$, making their usage steps, conditions, and constraints explicit for
comparison with guidance $M$. A language model derives this knowledge from the
full source code, assisted, where parsing is supported, by a syntax index of
definitions and parameters, call expressions, and conditional branches.\looseness=-1

Related implementation details are organized into workflow-level records
$\mathcal{K}=\{k_1,\ldots,k_m\}$. Each record connects what an operation
provides with the conditions and steps for using it:
\begin{equation}
  k=(\mathrm{cap},\ \mathrm{cond},\ \mathrm{steps},\ \mathrm{limits},\ \mathrm{src}),
  \label{eq:operational-knowledge}
\end{equation}
where the fields denote the capability, applicability conditions, ordered usage
steps, implementation constraints, and supporting source references,
respectively. This representation brings together information distributed
across the implementation into a description of how
\mbox{related operations can be used}.

For example, an object-counting record combines image and template inputs,
matching and deduplication parameters, and the use of printed output to obtain
the count. It therefore describes both how to invoke the operation and how to
access its result. The model associates each record with supporting source files
and line ranges; \system{} checks these locations and attaches the
\mbox{corresponding code}.

\Needspace{6\baselineskip}
\subsection{Constructing Workflow-Grounded Feedback}
\system{} compares operational knowledge $\mathcal{K}$ with guidance $M$, using
task request $x$ and historical execution trace $\tau$ as context. For
task-relevant workflows, the model checks whether $M$ specifies the necessary
operations, reflects their usage conditions, and provides an alternative step
when an implementation limitation applies.

The model organizes the comparisons into alignment records connecting task
requirements, existing instructions, supporting knowledge, and identified
guidance gaps. Records requiring changes produce feedback
$\mathcal{F}=\{f_1,\ldots,f_n\}$. Each item describes the issue, recommends a
change, and specifies when it applies, with references to the supporting
knowledge and task requirements. Corrections additionally cite the
\mbox{conflicting passages in $M$}.

In the object-counting example, $M$ gives general instructions to inspect scripts
and use runtime evidence, while $\mathcal{K}$ describes the tool's inputs and
count output. The resulting feedback addresses the missing invocation and
output-handling steps by recommending explicit invocation arguments and reading
the printed count for each \mbox{frame--template pair}.

\begin{table*}[t]
\centering\small
\setlength{\tabcolsep}{3.5pt}
\renewcommand{\arraystretch}{0.94}
\begin{tabular}{p{55pt}p{78pt}*{9}{>{\centering\arraybackslash}p{\dimexpr(\textwidth-133pt-22\tabcolsep)/9\relax}}}
\toprule
\multirow{2}{*}{Method} & \multirow{2}{*}{Condition}
 & \multicolumn{3}{c}{SkillsBench} & \multicolumn{3}{c}{SkillLearnBench}
 & \multicolumn{3}{c}{Overall} \\
\cmidrule(lr){3-5}\cmidrule(lr){6-8}\cmidrule(l){9-11}
 & & P@1 & P@3 & $\mathrm{Hit}^{3}$
   & P@1 & P@3 & $\mathrm{Hit}^{3}$
   & P@1 & P@3 & $\mathrm{Hit}^{3}$ \\
\midrule
Skill-Creator & --- & 39.68 & 53.57 & 25.00 & 39.39 & 54.55 & 24.24 & 39.53 & 54.10 & 24.59 \\
\midrule
\multirow{5}{*}{Self-Refine} & No-verifier & 37.30 & 51.19 & 22.62 & 40.74 & 58.59 & 25.25 & 39.16 & 55.19 & 24.04 \\
 & Binary & 38.49 & 53.57 & 22.62 & 36.36 & 50.51 & 21.21 & 37.34 & 51.91 & 21.86 \\
 & Report & 40.08 & 53.57 & 26.19 & 46.13 & \textbf{63.64} & 27.27 & 43.35 & 59.02 & 26.78 \\
 & Diagnostic & 45.63 & \textbf{63.10} & 28.57 & 47.81 & 61.62 & 33.33 & 46.81 & \textbf{62.30} & 31.15 \\
\rowcolor{gray!12}
 & VISE & \textbf{50.81} & 57.32 & \textbf{42.68} & \textbf{49.14} & 61.86 & \textbf{36.08} & \textbf{49.91} & 59.78 & \textbf{39.11} \\
\midrule
\multirow{5}{*}{TraceEvo} & No-verifier & 38.10 & 54.76 & 22.62 & 40.07 & 54.55 & 25.25 & 39.16 & 54.64 & 24.04 \\
 & Binary & 37.30 & 54.76 & 21.43 & 43.10 & 61.62 & 23.23 & 40.44 & 58.47 & 22.40 \\
 & Report & 39.29 & 50.00 & 27.38 & 43.43 & 61.62 & 27.27 & 41.53 & 56.28 & 27.32 \\
 & Diagnostic & 40.87 & \textbf{60.71} & 23.81 & 45.45 & 63.64 & 30.30 & 43.35 & 62.30 & 27.32 \\
\rowcolor{gray!12}
 & VISE & \textbf{52.03} & 58.54 & \textbf{43.90} & \textbf{52.92} & \textbf{65.98} & \textbf{37.11} & \textbf{52.51} & \textbf{62.57} & \textbf{40.22} \\
\midrule
\multirow{6}{*}{SkillLearn} & No-verifier & 38.10 & 53.57 & 22.62 & 39.06 & 55.56 & 23.23 & 38.62 & 54.64 & 22.95 \\
 & Binary & 36.90 & 47.62 & 23.81 & 43.43 & 60.61 & 24.24 & 40.44 & 54.64 & 24.04 \\
 & Report & 40.48 & 54.76 & 23.81 & 45.79 & \textbf{65.66} & 24.24 & 43.35 & 60.66 & 24.04 \\
 & Diagnostic & 38.49 & 55.95 & 21.43 & 39.73 & 52.53 & 28.28 & 39.16 & 54.10 & 25.14 \\
\rowcolor{gray!12}
 & VISE & \textbf{51.44} & \textbf{58.02} & \textbf{44.44} & \textbf{50.69} & 64.58 & \textbf{38.54} & \textbf{51.04} & \textbf{61.58} & \textbf{41.24} \\
\cmidrule(l){2-11}
 & Native & 51.41 & 55.42 & 45.78 & 52.53 & 56.57 & 46.46 & 52.01 & 56.04 & 46.15 \\
\midrule
\multirow{6}{*}{SkillRevise} & No-verifier & 43.25 & 59.52 & 27.38 & 41.41 & 56.57 & 29.29 & 42.26 & 57.92 & 28.42 \\
 & Binary & 43.25 & 53.57 & 33.33 & 42.42 & 59.60 & 28.28 & 42.81 & 56.83 & 30.60 \\
 & Report & 47.62 & \textbf{60.71} & 33.33 & \textbf{53.20} & \textbf{68.69} & 37.37 & 50.64 & \textbf{65.03} & 35.52 \\
 & Diagnostic & 50.79 & \textbf{60.71} & 39.29 & 48.82 & 63.64 & 35.35 & 49.73 & 62.30 & 37.16 \\
\rowcolor{gray!12}
 & VISE & \textbf{51.22} & 58.54 & \textbf{42.68} & 51.55 & 67.01 & \textbf{39.18} & \textbf{51.40} & 63.13 & \textbf{40.78} \\
\cmidrule(l){2-11}
 & Native & 50.20 & 56.63 & 42.17 & 57.91 & 64.65 & 47.47 & 54.40 & 60.99 & 45.05 \\
\bottomrule

\end{tabular}
\caption{Downstream success (\%). Bold marks column-wise best scores among the five
feedback conditions within each method; gray rows denote VISE. Native reports each
method's original workflow for comparison. Overall weights tasks equally. See
Appendix~\ref{app:main_protocol} for task coverage and evaluation protocols.}
\label{tab:main}
\end{table*}

\subsection{Revising Workflow Guidance}
Given task request $x$, operational knowledge $\mathcal{K}$, and feedback
$\mathcal{F}$, the revision model generates targeted edits to guidance $M$ while
preserving useful existing procedures. Each edit pairs an original passage with
replacement instructions and references the feedback items it addresses.

\system{} checks that the original passages match unique locations in $M$ and
that the edits do not overlap. It then applies the replacements to obtain $M'$,
leaving scripts $P$ unchanged:
\begin{equation}
  S=(P,M)\;\longrightarrow\;S'=(P,M').
  \label{eq:guidance-revision}
\end{equation}
{\widowpenalty=10000\clubpenalty=10000
The revised skill $S'$ is frozen before downstream execution and task-verifier
evaluation. Editing limits and validation details are provided in
Appendix~\ref{app:implementation_details}.\par}

\section{Experiments}
\label{sec:experiments}

We evaluate \system{} on SkillsBench and SkillLearnBench to examine whether
tool-grounded feedback improves skills without using task-verifier judgments
during revision. Across
four evolution methods, we compare VISE-refined skills with the corresponding
No-verifier skills and skills evolved with task-verifier feedback. We measure
average success, task coverage, and repeated success to characterize their
downstream utility.

\Needspace{7\baselineskip}
\subsection{Experimental Setup}

\paragraph{Tasks and initial skills.}
We evaluate on SkillsBench and SkillLearnBench~\citep{li2026skillsbench,zhong2026skilllearnbench}.
For each task, we start skill evolution from an initial skill $S_0=(P,M_0)$, comprising public
executable scripts $P$ and initial workflow guidance $M_0$. We include
Skill-Creator~\citep{anthropic2026skillcreator} as the unevolved baseline.

\paragraph{Evolution methods.}
We evaluate four skill self-evolution methods. The Self-Refine
baseline~\citep{madaan2023selfrefine} directly revises
workflow guidance $M$ in \skillmd{} using the available inputs. TraceEvo reflects on execution traces
to identify guidance deficiencies.
SkillLearn~\citep{zhong2026skilllearnbench} examines the task request and existing skill for missing or inaccurate
instructions. SkillRevise~\citep{liu2026skillrevise} diagnoses skill defects and retrieves repair principles
to guide revision. For SkillLearn and SkillRevise, we also evaluate
their native workflows with at most one revision, labeled Native in
Table~\ref{tab:main}.
We use GPT-5.5 for both skill evolution and downstream execution.\looseness=-1

\paragraph{Feedback conditions.}
All conditions use the task request $x$, scripts $P$, workflow guidance $M$,
and historical execution trace $\tau$. No-verifier uses these inputs alone.
Binary adds the verifier's
PASS/FAIL judgment on the initial execution. Report additionally provides
test-failure reports, including expected and observed values when available.
Diagnostic provides the judgment together with native evaluation metadata,
diagnostic messages, and failure traces. VISE provides structural feedback $\mathcal{F}$ by
aligning operational knowledge $\mathcal{K}$ extracted from $P$ with guidance
$M$, using $x$ and $\tau$ as context. Scripts remain fixed during guidance revision.

\paragraph{Evaluation.}
For each task, we freeze the final skill and evaluate it over three downstream
executions using the benchmark's native verifier. Following the distinction between
task coverage and repeated reliability~\citep{yao2024taubench}, we report P@1 as the average
success rate, P@3 as the proportion of tasks succeeding at least once, and
$\mathrm{Hit}^{3}$ as the proportion of tasks succeeding in all three executions.
Results are reported for each benchmark and pooled into Overall with equal
weight per task. Evaluation protocols and task coverage are detailed in
Appendix~\ref{app:main_protocol}.

\begin{figure*}[t]
\centering
\includegraphics[width=\textwidth]{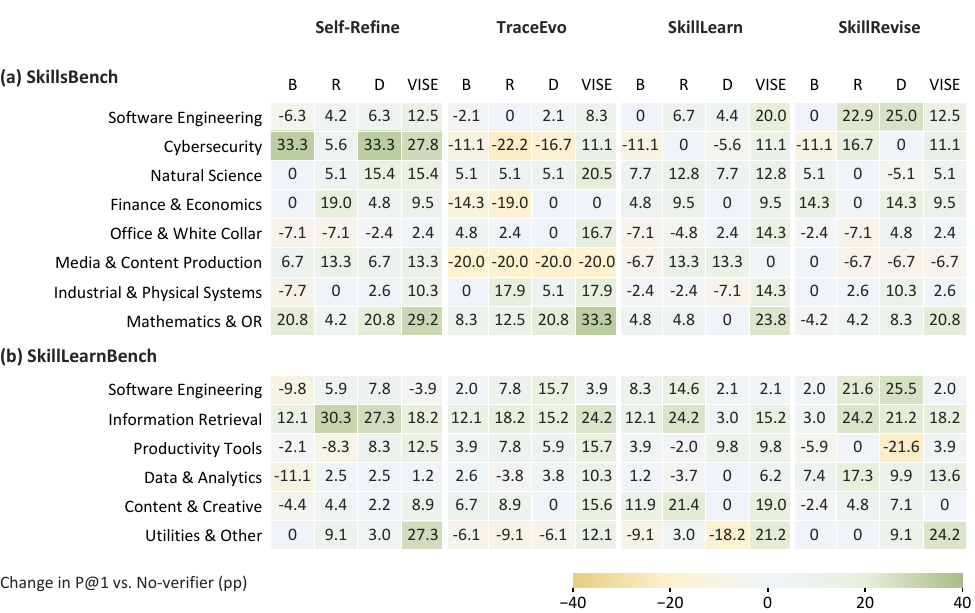}
\caption{Feedback gains vary by task category and evolution method.
Cells show P@1 changes (percentage points) from Table~\ref{tab:main}'s
No-verifier condition on each method's common complete-task set.
B/R/D: Binary/Report/Diagnostic.}
\label{fig:verifier_category_effects}
\end{figure*}

\Needspace{6\baselineskip}
\subsection{Main Results}
\label{sec:results}

\textbf{Feedback effectiveness varies across evolution methods.}
As shown in Table~\ref{tab:main}, only SkillRevise exceeds Skill-Creator in
Overall P@1 under No-verifier. Relative to No-verifier, Binary reduces
Self-Refine's Overall P@1, while Report improves it across all four methods by
\mbox{2.37--8.38\%}. Switching from Report to Diagnostic increases Overall P@1 for
Self-Refine and TraceEvo but decreases it for SkillLearn and SkillRevise.
VISE achieves higher Overall P@1, P@3, and $\mathrm{Hit}^{3}$ than No-verifier
across all four evolution methods, with gains of \mbox{9.14--13.35\%}, \mbox{4.59--7.93\%},
and \mbox{12.36--18.29\%}, respectively.

\textbf{VISE achieves higher Overall P@1 and $\mathrm{Hit}^{3}$ than the three
task-verifier feedback conditions.} As shown in Table~\ref{tab:main}, VISE exceeds
the best task-verifier condition for each metric by \mbox{0.76--9.16\%} in P@1 and
\mbox{3.62--16.10\%} in $\mathrm{Hit}^{3}$, with a larger margin in $\mathrm{Hit}^{3}$
for every evolution method. Overall P@3 rankings vary across methods, with VISE
achieving the highest scores for TraceEvo and SkillLearn. For the native workflows
of SkillLearn and SkillRevise, the comparison differs: Native achieves higher
Overall P@1 and $\mathrm{Hit}^{3}$, while VISE achieves higher Overall~P@3.


\section{Analysis}
\label{sec:rq_analysis}

\subsection{RQ1: Feedback Effects across Methods and Task Categories}
\label{sec:rq1_verifier}
\textbf{The relative effectiveness of VISE and task-verifier feedback varies
across task categories.}
As shown in Figure~\ref{fig:verifier_category_effects}, VISE achieves higher P@1
than the three task-verifier feedback conditions on Mathematics \& OR in SkillsBench and
Utilities \& Other in SkillLearnBench across all four evolution methods.
The VISE--Report ordering reverses on Software Engineering in SkillLearnBench:
Report gains 5.9--21.6\% over No-verifier, while VISE yields
changes of $-3.9\%$ to 3.9\%.

\textbf{Within the same task category, feedback gains also depend on the
evolution method.} On Productivity Tools in SkillLearnBench,
Diagnostic improves P@1 over No-verifier for Self-Refine, TraceEvo, and SkillLearn,
but lowers P@1 by 21.6\% for SkillRevise. By contrast, VISE
improves P@1 over No-verifier by 3.9--15.7\% across all four methods in this
category.


\subsection{RQ2: Skill Improvement without Task-Verifier Feedback}
\label{sec:rq2_consistency}
\begin{figure*}[t]
\centering
\includegraphics[width=\textwidth]{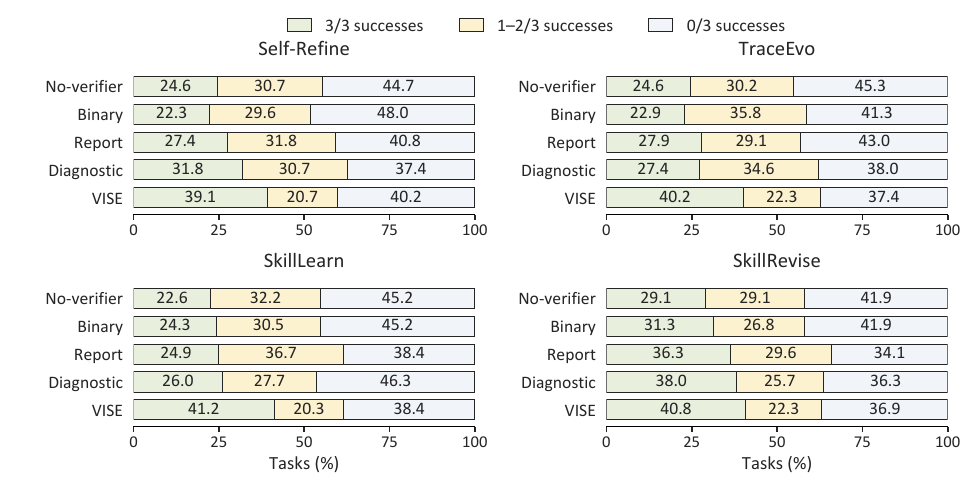}
\caption{Success-count distributions across feedback conditions.
Bars show task proportions with three, one or two, and zero successes across
three evaluation records. All five conditions use the same tasks within each method.}
\label{fig:repeated_success}
\end{figure*}

VISE shifts the success distribution toward tasks solved in all three
executions. As shown in Figure~\ref{fig:repeated_success}, tasks with only one or
two successful executions account for 20.3\%--22.3\% under VISE, compared with
29.1\%--32.2\% under No-verifier. On the common SkillLearn task set, VISE and
Report both achieve P@3 of 61.58\%, but 67.0\% of tasks solved at least once
succeed in all three executions under VISE, compared with 40.4\% under Report.
Their equal task coverage thus conceals different repeated-success patterns.

\begin{figure}[!t]
\centering
\setlength{\abovecaptionskip}{4pt}
\includegraphics[width=0.92\columnwidth]{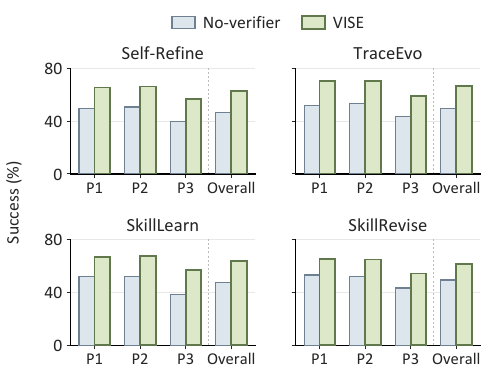}
\caption{VISE improves transfer to task variants.
P1/P2/P3: representation, instance, and coupled-requirement changes.}
\label{fig:variant_transfer}
\vspace{-12pt}
\end{figure}

The revised skills also retain an advantage when transferred to task variants
without further revision. We freeze the skills revised on the original tasks
and evaluate them directly on the 233 SkillsBench variants from
Section~\ref{sec:task_variants}. As shown in Figure~\ref{fig:variant_transfer}, VISE improves success
rates in each of the three variant categories across all four methods. Overall
success rises from \mbox{46.35\%--49.36\%} to \mbox{61.37\%--66.52\%}.
Coupled-requirement variants
remain the lowest-scoring category for every method, even after revision.

\subsection{RQ3: Implementation Evidence in Guidance and Tool Use}
\label{sec:rq3_implementation}
We revisit the SkillLearnBench video object-counting example from
Section~\ref{sec:method} to examine how revised guidance is reflected in
execution. VISE expands the Self-Refine skill's general instruction to inspect
scripts into a counting procedure that specifies how to invoke the tool and use
its printed output for each frame--template pair.

Both executions invoke the counter, but integrate it differently into the
workflow. In the original execution, the recorded calls process a single frame
after the CSV has already been created. With the revised skill, the agent invokes
the counter within a frame--template loop, parses its printed counts, and
aggregates them into the final CSV. This execution follows the added output-use
instruction: the counter's results become inputs to the \mbox{final result construction}.

\subsection{Component Comparisons}
\label{sec:component_comparisons}
\begin{table}[!htbp]
\centering
\small
\setlength{\tabcolsep}{3pt}
\renewcommand{\arraystretch}{1.00}
\begin{tabular*}{\linewidth}{@{\extracolsep{\fill}}lccc@{}}
\toprule
\textbf{Condition} & \textbf{P@1} & \textbf{P@3} & $\boldsymbol{\mathrm{Hit}^{3}}$ \\
\midrule
 w/o Workflow Revision & $\downarrow$8.33 & 0.00 & $\downarrow$5.00 \\
\addlinespace[1pt]
w/o Structured Feedback & $\downarrow$13.89 & $\downarrow$16.67 & $\downarrow$5.00 \\
\addlinespace[1pt]
w/o Workflow Alignment &
$\downarrow$10.56 & $\downarrow$5.00 & $\downarrow$1.67 \\
\midrule
Raw-Source Revision & $\downarrow$18.89 & $\downarrow$13.33 & $\downarrow$16.67 \\
\midrule
Full VISE & 57.78 & 76.67 & 30.00 \\
\bottomrule
\end{tabular*}
\caption{TraceEvo component comparisons on 60 tasks. Full VISE reports
absolute scores (\%); other rows report percentage-point decreases from it.}
\label{tab:component_ablation}
\end{table}

We select 60 tasks that come with scripts in the original benchmarks.
All component comparisons use TraceEvo.
Full VISE achieves higher P@1 and
$\mathrm{Hit}^{3}$ than w/o Workflow Revision, with unchanged P@3, as shown in
Table~\ref{tab:component_ablation}. Among tasks solved at least once, the mean
success count increases from 1.93 to 2.26: the higher P@1 reflects more repeated
successes rather than broader task coverage. Full VISE also solves 46 tasks at least once,
compared with 36 under direct source-and-trace revision without structured
feedback.

Compared with w/o Workflow Alignment, Full VISE has a larger advantage in P@1
than in $\mathrm{Hit}^{3}$, with fewer one-success tasks, more two-success tasks,
and similar three-success counts. Condition definitions and complete
success-count distributions appear in Appendix~\ref{app:ablation_protocol}.


\section{Conclusion}
\system{} grounds skill evolution in executable tools: scripts establish the
available operations, and their implementations provide evidence for revising
workflow guidance. Our empirical study shows that similar aggregate gains can
involve different benefiting tasks, while changes in task demands can weaken or
reverse skill benefits. Across four evolution methods on SkillsBench and
SkillLearnBench, \system{} improves average success, task coverage, and three-run
success over the corresponding No-verifier conditions. These results suggest that
verifier-independent skill evolution can benefit from explicitly connecting what
tools can do with how workflow guidance directs their use.

\section*{Limitations}
\system{} derives revision feedback from inspectable tool implementations.
For skills that rely on closed-source tools or remote APIs, accessible scripts
may expose invocation logic but provide limited information about the underlying
operations. This restricts the operational knowledge available for workflow
revision. Incorporating tool documentation and observed execution behavior is a
natural direction for extending \system{} to these settings.

\section*{Ethical Considerations}
Revised skill guidance may be reused across tasks, allowing unsafe instructions or sensitive
information to propagate beyond the original interaction. Deployments should retain tool-level
access controls and review updates before reuse in security-sensitive settings. Shared artifacts
should preserve source attribution and exclude credentials and private task data.

\bibliography{references}

@article{wang2023voyager,
  title     = {Voyager: An Open-Ended Embodied Agent with Large Language Models},
  author    = {Wang, Guanzhi and Xie, Yuqi and Jiang, Yunfan and Mandlekar, Ajay and Xiao, Chaowei and Zhu, Yuke and Fan, Linxi and Anandkumar, Anima},
  journal   = {arXiv preprint arXiv:2305.16291},
  year      = {2023},
  doi       = {10.48550/arXiv.2305.16291},
  url       = {https://arxiv.org/abs/2305.16291}
}

@article{li2026skillsbench,
  title     = {SkillsBench: Benchmarking How Well Agent Skills Work Across Diverse Tasks},
  author    = {Li, Xiangyi and Liu, Yimin and Chen, Wenbo and You, Bingran and Di, Zonglin and He, Yifeng and Zheng, Shenghan and others},
  journal   = {arXiv preprint arXiv:2602.12670},
  year      = {2026},
  doi       = {10.48550/arXiv.2602.12670},
  note      = {Version 4},
  url       = {https://arxiv.org/abs/2602.12670v4}
}

@article{liu2026skillrevise,
  title = {{SkillRevise}: Improving {LLM}-Authored Agent Skills via Trace-Conditioned Skill Revision},
  author = {Liu, Yuxuan and Su, Zhaochen and Xie, Lingyun and Zhang, Yuhao and Zong, Qing and Guo, Jiahe and Xie, Zhongwei and Ji, Yiyan and Yim, Yauwai and Luo, Hongyu and Ren, Xiyu and Ruan, Chenyu and Li, Haoran and Song, Yangqiu},
  journal = {arXiv preprint arXiv:2606.01139},
  year = {2026},
  note = {Version 4},
  doi = {10.48550/arXiv.2606.01139},
  url = {https://arxiv.org/abs/2606.01139v4}
}

@article{zhang2026skillevolver,
  title = {{SkillEvolver}: Skill Learning as a Meta-Skill},
  author = {Zhang, Genrui and Zhu, Erle and Zhou, Jinfeng and Jia, Caiyan and Wang, Hongning},
  journal = {arXiv preprint arXiv:2605.10500},
  year = {2026},
  doi = {10.48550/arXiv.2605.10500},
  note = {Version 1},
  url = {https://arxiv.org/abs/2605.10500v1}
}

@article{zhou2026spark,
  title = {Evidence Over Plans: Online Trajectory Verification for Skill Distillation},
  author = {Zhou, Yang and Dong, Zihan and Wang, Zhenting and Jin, Can and Zhao, Shiyu and Guo, Bangwei and Gu, Difei and Zhang, Linjun and Zhou, Mu and Metaxas, Dimitris N.},
  journal = {arXiv preprint arXiv:2605.09192},
  year = {2026},
  doi = {10.48550/arXiv.2605.09192},
  note = {Version 2},
  url = {https://arxiv.org/abs/2605.09192v2}
}

@article{chen2024automanual,
  title = {{AutoManual}: Constructing Instruction Manuals by {LLM} Agents via Interactive Environmental Learning},
  author = {Chen, Minghao and Li, Yihang and Yang, Yanting and Yu, Shiyu and Lin, Binbin and He, Xiaofei},
  journal = {arXiv preprint arXiv:2405.16247},
  year = {2024},
  note = {Version 4},
  doi = {10.48550/arXiv.2405.16247},
  url = {https://arxiv.org/abs/2405.16247v4}
}

@inproceedings{qu2025draft,
  title     = {From Exploration to Mastery: Enabling {LLMs} to Master Tools via Self-Driven Interactions},
  author    = {Qu, Changle and Dai, Sunhao and Wei, Xiaochi and Cai, Hengyi and Wang, Shuaiqiang and Yin, Dawei and Xu, Jun and Wen, Ji-Rong},
  booktitle = {International Conference on Learning Representations},
  year      = {2025},
  doi       = {10.48550/arXiv.2410.08197},
  url       = {https://arxiv.org/abs/2410.08197v2}
}

@article{lu2026docschisel,
  title     = {{DocsChisel}: Adaptive Tool Documentation Optimization Framework for {LLM} Agents},
  author    = {Lu, You and Zhang, Kun and Chen, Bihuan and Peng, Xin},
  journal   = {arXiv preprint arXiv:2608.10037},
  year      = {2026},
  note      = {Version 1},
  doi       = {10.48550/arXiv.2608.10037},
  url       = {https://arxiv.org/abs/2608.10037v1}
}

@article{fang2025play2prompt,
  title = {{Play2Prompt}: Zero-shot Tool Instruction Optimization for {LLM} Agents via Tool Play},
  author = {Fang, Wei and Zhang, Yang and Qian, Kaizhi and Glass, James and Zhu, Yada},
  journal = {arXiv preprint arXiv:2503.14432},
  year = {2025},
  note = {Version 2},
  doi = {10.48550/arXiv.2503.14432},
  url = {https://arxiv.org/abs/2503.14432v2}
}

@article{zhong2026skilllearnbench,
  title     = {SkillLearnBench: Benchmarking Continual Learning Methods for Agent Skill Generation on Real-World Tasks},
  author    = {Zhong, Shanshan and Lu, Yi and Ning, Jingjie and Wan, Yibing and Feng, Lihan and Ao, Yuyi and Ribeiro, Leonardo F. R. and Dreyer, Markus and Ammirati, Sean and Xiong, Chenyan},
  journal   = {arXiv preprint arXiv:2604.20087},
  year      = {2026},
  doi       = {10.48550/arXiv.2604.20087},
  note      = {Version 1},
  url       = {https://arxiv.org/abs/2604.20087v1}
}

@article{zhang2026coevoskills,
  title     = {CoEvoSkills: Self-Evolving Agent Skills via Co-Evolutionary Verification},
  author    = {Zhang, Hanrong and Fan, Shicheng and Zou, Henry Peng and Chen, Yankai and Wang, Zhenting and Zhou, Jiayu and Li, Chengze and Huang, Wei-Chieh and Yao, Yifei and Zheng, Kening and Liu, Xue and Li, Xiaoxiao and Yu, Philip S.},
  journal   = {arXiv preprint arXiv:2604.01687},
  year      = {2026},
  note      = {Version 3},
  doi       = {10.48550/arXiv.2604.01687},
  url       = {https://arxiv.org/abs/2604.01687v3}
}

@article{alzubi2026evoskill,
  title     = {EvoSkill: Automated Skill Discovery for Multi-Agent Systems},
  author    = {Alzubi, Salaheddin and Provenzano, Noah and Bingham, Jaydon and Chen, Weiyuan and Vu, Tu},
  journal   = {arXiv preprint arXiv:2603.02766},
  year      = {2026},
  note      = {Version 1},
  doi       = {10.48550/arXiv.2603.02766},
  url       = {https://arxiv.org/abs/2603.02766v1}
}

@article{yang2026skillopt,
  title     = {SkillOpt: Executive Strategy for Self-Evolving Agent Skills},
  author    = {Yang, Yifan and Gong, Ziyang and Huang, Weiquan and Yang, Qihao and Zhou, Ziwei and Huang, Zisu and Li, Yan and Gao, Xuemei and Dai, Qi and Liu, Bei and Qiu, Kai and Yang, Yuqing and Chen, Dongdong and Yang, Xue and Luo, Chong},
  journal   = {arXiv preprint arXiv:2605.23904},
  year      = {2026},
  note      = {Version 2},
  doi       = {10.48550/arXiv.2605.23904},
  url       = {https://arxiv.org/abs/2605.23904v2}
}

@inproceedings{madaan2023selfrefine,
  title     = {Self-Refine: Iterative Refinement with Self-Feedback},
  author    = {Madaan, Aman and Tandon, Niket and Gupta, Prakhar and Hallinan, Skyler and Gao, Luyu and Wiegreffe, Sarah and Alon, Uri and Dziri, Nouha and Prabhumoye, Shrimai and Yang, Yiming and Gupta, Shashank and Majumder, Bodhisattwa Prasad and Hermann, Katherine and Welleck, Sean and Yazdanbakhsh, Amir and Clark, Peter},
  booktitle = {Advances in Neural Information Processing Systems},
  volume    = {36},
  year      = {2023},
  doi       = {10.52202/075280-2019},
  url       = {https://proceedings.neurips.cc/paper_files/paper/2023/hash/91edff07232fb1b55a505a9e9f6c0ff3-Abstract-Conference.html}
}

@inproceedings{shinn2023reflexion,
  title     = {Reflexion: Language Agents with Verbal Reinforcement Learning},
  author    = {Shinn, Noah and Cassano, Federico and Gopinath, Ashwin and Narasimhan, Karthik and Yao, Shunyu},
  booktitle = {Advances in Neural Information Processing Systems},
  volume    = {36},
  year      = {2023},
  doi       = {10.52202/075280-0377},
  url       = {https://proceedings.neurips.cc/paper_files/paper/2023/hash/1b44b878bb782e6954cd888628510e90-Abstract-Conference.html}
}

@inproceedings{cai2024latm,
  title     = {Large Language Models as Tool Makers},
  author    = {Cai, Tianle and Wang, Xuezhi and Ma, Tengyu and Chen, Xinyun and Zhou, Denny},
  booktitle = {International Conference on Learning Representations},
  year      = {2024},
  url       = {https://proceedings.iclr.cc/paper_files/paper/2024/hash/ed91353f700d113e5d848c7e04a858b0-Abstract-Conference.html}
}

@inproceedings{qian2023creator,
  title     = {{CREATOR}: Tool Creation for Disentangling Abstract and Concrete Reasoning of Large Language Models},
  author    = {Qian, Cheng and Han, Chi and Fung, Yi and Qin, Yujia and Liu, Zhiyuan and Ji, Heng},
  booktitle = {Findings of the Association for Computational Linguistics: EMNLP 2023},
  pages     = {6922--6939},
  publisher = {Association for Computational Linguistics},
  year      = {2023},
  doi       = {10.18653/v1/2023.findings-emnlp.462},
  url       = {https://aclanthology.org/2023.findings-emnlp.462/}
}

@article{yao2024taubench,
  title     = {{$\tau$}-bench: A Benchmark for Tool-Agent-User Interaction in Real-World Domains},
  author    = {Yao, Shunyu and Shinn, Noah and Razavi, Pedram and Narasimhan, Karthik},
  journal   = {arXiv preprint arXiv:2406.12045},
  year      = {2024},
  note      = {Version 1},
  doi       = {10.48550/arXiv.2406.12045},
  url       = {https://arxiv.org/abs/2406.12045v1}
}

@article{chen2026repotoskill,
  title = {{Repo-To-Skill}: Distilling {GitHub} Repositories Into {AI4AI} Skills},
  author = {Chen, Jianlyu and Hu, Yuyang and Qian, Hongjin and Liu, Jiawei and Wei, Wenqing and Chen, Xiaolong and Lian, Defu and Dou, Zhicheng and Li, Chaozhuo and Ye, Qiwei and Liu, Zheng},
  journal = {arXiv preprint arXiv:2609.02749},
  year = {2026},
  note = {Version 1},
  doi = {10.48550/arXiv.2609.02749},
  url = {https://arxiv.org/abs/2609.02749v1}
}

@article{liu2026skillscriptbench,
  title     = {SkillScriptBench: Benchmarking Self-Evolution of Executable Agent Skill Packages Beyond Markdown},
  author    = {Liu, Yuxuan and Li, Haoran and Zhang, Yuhao and Guo, Jiahe and Luo, Hongyu and Hu, Wenbin and Jing, Huihao and Chung, Kawai and Chen, Junle and Fan, Changxuan and Zong, Qing and Xie, Lingyun and Song, Yangqiu},
  journal   = {arXiv preprint arXiv:2610.04008},
  year      = {2026},
  doi       = {10.48550/arXiv.2610.04008},
  url       = {https://arxiv.org/abs/2610.04008}
}

@article{liu2026rethinking,
  title = {Rethinking Self-Evolving Agent Skills: Feedback Dynamics over Multiple Rounds},
  author = {Liu, Yuxuan and Su, Zhaochen and Zhang, Yuhao and Guo, Jiahe and Xie, Zhongwei and Jing, Huihao and Xie, Lingyun and Zong, Qing and Yim, Yauwai and Zhang, Zhixiong and Li, Haoran and Song, Yangqiu},
  journal = {arXiv preprint arXiv:2608.02636},
  year = {2026},
  doi = {10.48550/arXiv.2608.02636},
  url = {https://arxiv.org/abs/2608.02636}
}

@article{liu2026skillhorizon,
  title = {{SkillHorizon}: Constructing Long-Horizon Agent Tasks through Skill Composition},
  author = {Liu, Yuxuan and Su, Zhaochen and Guo, Hangyu and Huang, Shijue and Zhang, Zhixiong and Chung, Kawai and Xie, Zhongwei and Zheng, Tianshi and Fan, Wei and Zong, Qing and Shi, Haochen and Xu, Baixuan and Li, Haoran and Song, Yangqiu},
  journal = {arXiv preprint arXiv:2610.14966},
  year = {2026},
  doi = {10.48550/arXiv.2610.14966},
  url = {https://arxiv.org/abs/2610.14966}
}

@misc{zhang2025agentskills,
  title = {Equipping Agents for the Real World with {Agent Skills}},
  author = {Zhang, Barry and Lazuka, Keith and Murag, Mahesh},
  year = {2025},
  month = oct,
  howpublished = {Anthropic Engineering},
  url = {https://www.anthropic.com/engineering/equipping-agents-for-the-real-world-with-agent-skills}
}

@misc{anthropic2026skillcreator,
  title = {{Skill-Creator}},
  author = {{Anthropic}},
  year = {2026},
  howpublished = {Software documentation},
  note = {Accessed October 13, 2026},
  url = {https://github.com/anthropics/skills/blob/main/skills/skill-creator/SKILL.md}
}
\clearpage
\appendix
\twocolumn[{
\section{Diagnostic Study: Data and Analysis}
\label{app:diagnostic}

\begin{minipage}{\textwidth}
\centering\small
\setlength{\tabcolsep}{3pt}
\begin{tabular*}{\textwidth}{@{\extracolsep{\fill}}lrrrrrrrr@{}}
\toprule
\multirow{2}{*}{Method} & \multicolumn{4}{c}{Original tasks}
 & \multicolumn{4}{c}{Task variants} \\
\cmidrule(lr){2-5}\cmidrule(l){6-9}
 & Eval. 1 & Eval. 2 & Mean & $|\Delta|$ & Representation & Instance & Requirements & Overall \\
\midrule
\multicolumn{9}{@{}l}{\textit{Shared references}} \\
No-skill & 45.74 & 37.66 & 41.70 & 8.08 & 40.56 & 44.66 & 38.50 & 41.09 \\
Skill-Creator & 42.15 & 33.24 & 37.70 & 8.91 & 47.96 & 39.44 & 37.58 & 41.76 \\
\midrule
\multicolumn{9}{@{}l}{\textit{Evolved skills}} \\
SkillEvolver & 63.22 & 62.34 & 62.78 & 0.87 & 53.89 & 59.56 & 48.65 & 53.81 \\
SPARK & 36.63 & 40.93 & 38.78 & 4.30 & 44.05 & 39.77 & 42.66 & 42.26 \\
CoEvoSkills & 52.44 & 46.44 & 49.44 & 6.00 & 38.46 & 35.49 & 35.71 & 36.60 \\
Self-Refine & 60.38 & 60.54 & 60.46 & 0.16 & 60.21 & 63.08 & 52.92 & 58.55 \\
SkillLearn & 54.43 & 55.80 & 55.11 & 1.36 & 64.50 & 58.88 & 50.14 & 57.80 \\
Trace Evo & 63.76 & 59.89 & 61.82 & 3.87 & 62.11 & 64.39 & 50.69 & 58.84 \\
SkillRevise & 57.71 & 57.43 & 57.57 & 0.28 & 58.27 & 54.88 & 50.01 & 54.36 \\
\bottomrule
\end{tabular*}
\captionof{table}{Detailed mean rewards for original-task evaluations and variant categories (0--100).
$|\Delta|$ is the absolute difference between Eval.~1 and Eval.~2, computed before rounding.
Original-task means weight the two evaluations equally.}
\label{tab:motivation_detailed}
\end{minipage}

\vspace{12pt}
}]
\raggedbottom
\widowpenalty=10000
\clubpenalty=10000

\paragraph{Original-task evaluations.}
Table~\ref{tab:motivation_detailed} lists the individual evaluation scores
and category-level variant results.
Table~\ref{tab:motivation_combined} averages two evaluations for
SkillEvolver (81 tasks), SPARK (84 tasks), CoEvoSkills (85 tasks), and Self-Refine, SkillLearn, and
Trace Evo (86 tasks each). Skill-Creator averages two evaluations
on the same 86 tasks. SkillRevise uses 86 and 83 tasks in its two evaluations.
Partial rewards retain their values, and budget-exhausted executions are scored
zero. All rows report mean reward on the 0--100 scale.

\paragraph{No-skill reference.}
The displayed original-task No-skill reference uses CoEvoSkills' two
85-task evaluations.

\subsection{Sources and recurrence of task-level gains}
\label{app:gain_recurrence}
Task identities match across the two evaluations within each method:
81 for SkillEvolver and 85 for CoEvoSkills.
Each evaluation pairs the evolved-skill condition with a No-skill execution of
the same task. For reward difference $d_i$ on the 0--100 scale, contributions are
$G=\sum_i\max(d_i,0)/N$ and $L=\sum_i\max(-d_i,0)/N$; their difference
is the paired mean gain. All $N$ tasks, including ties, enter the denominator.

\begin{table}[!htbp]
\centering\small
\begin{tabular*}{\columnwidth}{@{\extracolsep{\fill}}lrr@{}}
\toprule
Measure & Run 1 & Run 2 \\
\midrule
No-skill reward & 37.60 & 37.32 \\
Evolved reward & 63.22 & 62.34 \\
Positive contribution $G$ & 30.91 & 28.32 \\
Negative contribution $L$ & 5.30 & 3.30 \\
Net gain & 25.61 & 25.02 \\
Improved tasks & 27 & 26 \\
Tied tasks & 48 & 52 \\
Regressed tasks & 6 & 3 \\
\bottomrule
\end{tabular*}
\caption{Task-level gain decomposition for two SkillEvolver evaluations.}
\label{tab:gain_recurrence}
\end{table}

Of 36 SkillEvolver tasks improving in at least one evaluation, 17 improve in both;
the overlap is 17/36, whereas retention among first-evaluation improvements is
17/27. CoEvoSkills has 17 and 18 improved tasks, with 10 shared out of 25
improving in either evaluation. Table~\ref{tab:gain_transitions} shows the
transitions between evaluations. CoEvoSkills' first/second positive contributions
are 17.14/19.44, negative contributions 6.90/9.48, and net gains 10.24/9.96.

\begin{table}[!htbp]
\centering\small
\setlength{\tabcolsep}{3pt}
\begin{tabular*}{\columnwidth}{@{\extracolsep{\fill}}lrrrrrr@{}}
\toprule
& \multicolumn{3}{c}{SkillEvolver} & \multicolumn{3}{c}{CoEvoSkills} \\
\cmidrule(lr){2-4}\cmidrule(l){5-7}
Eval. 1 & I & T & R & I & T & R \\
\midrule
Improved & 17 & 10 & 0 & 10 & 7 & 0 \\
Tied & 7 & 38 & 3 & 8 & 47 & 4 \\
Regressed & 2 & 4 & 0 & 0 & 3 & 6 \\
\bottomrule
\end{tabular*}
\caption{Task-level gain transitions from Eval.~1 (rows) to Eval.~2
(columns). I/T/R denote improved, tied, and regressed relative to each
evaluation's paired No-skill result.}
\label{tab:gain_transitions}
\end{table}

\paragraph{Changes in execution outcomes.}
Self-Refine receives full reward on 50 tasks in
each evaluation, with nine tasks losing and nine gaining full reward.
SkillLearn's full-reward count changes from 44 to 46, with seven losses and nine
gains; Trace Evo's changes from 53 to 48, with ten losses and five gains.

\subsection{Task-variant construction}
\label{app:variant_construction}
The 233-variant collection derives from 81 original SkillsBench tasks:
81 representation variants (P1), 71 instance variants (P2), and 81
coupled-requirement variants (P3). Each variant retains its source-task identity
and includes a public request, task assets and environment configuration, and
a corresponding evaluation specification. The three categories distinguish
the intended change to the task, rather than prescribing one transformation
for every domain.

SkillHorizon~\citep{liu2026skillhorizon} studies a complementary task-construction
setting, composing reusable skill procedures into long-horizon tasks with
interdependent activities. Here, variants modify the requirements of existing
SkillsBench tasks.

\paragraph{Representation changes (P1).}
These variants change how an operation receives inputs or presents outputs,
including paths, field names, and serialization conventions. They test whether
a skill can recover the required operation from a changed representation
instead of relying on the original surface form.

\paragraph{Instance changes (P2).}
These variants replace concrete inputs or relations on which the workflow
operates, such as data records or authorization relationships. For example,
the dependency-audit variant changes package-location and dependency metadata
in its supplied lockfile. The task still requires dependency auditing, but the
workflow must derive its findings from the new instance.

\paragraph{Coupled-requirement changes (P3).}
These variants add or modify requirements that must be satisfied jointly.
In the 3D-scan task, the original request selects the largest connected
component and computes its mass from volume and material density. Its P3
variant selects the largest-volume component only among those with a material
in the supplied density table and a bounding-box centre whose X coordinate
lies in $[40,60]$ centimetres. The output retains the mass and material-ID
fields, but eligibility filtering must now precede component selection.

\paragraph{Packaging and evaluation.}
Variant requests specify the applicable input and output contracts, including
shared clarifications where needed. Public task assets and private grading
assets are packaged separately.
Skills come from the corresponding original SkillsBench tasks: generated skills
for Skill-Creator and evolved skills for the evolution methods. They remain
frozen during variant evaluation. No-skill supplies no skill. Each execution is
scored with the corresponding variant's grader.

\subsection{Utility by task-change category}
\label{app:variant_gains}
Table~\ref{tab:variant_gains} reports mean within-variant reward differences
between skill-assisted and No-skill executions on the 0--100 scale, using
each method's paired task cohort. Skill-Creator, Self-Refine, SkillLearn,
Trace Evo, and SkillRevise each cover 233 variants (81/71/81 for P1/P2/P3).
SkillEvolver covers 222 (77/68/77), while CoEvoSkills and SPARK each cover
221 (77/68/76). The paired No-skill means for SkillEvolver and CoEvoSkills
are 41.87 and 41.61, respectively. SPARK's overall score weights category
means by task count.

\begin{table}[!htbp]
\centering\small
\begin{tabular*}{\columnwidth}{@{\extracolsep{\fill}}lrrr@{}}
\toprule
Method & P1 & P2 & P3 \\
\midrule
Skill-Creator & $+7.40$ & $-5.22$ & $-0.92$ \\
SkillEvolver & $+11.85$ & $+14.56$ & $+9.72$ \\
SPARK & $+2.00$ & $-5.23$ & $+4.53$ \\
CoEvoSkills & $-3.59$ & $-9.51$ & $-2.42$ \\
Self-Refine & $+19.66$ & $+18.42$ & $+14.41$ \\
SkillLearn & $+23.95$ & $+14.23$ & $+11.64$ \\
Trace Evo & $+21.56$ & $+19.73$ & $+12.19$ \\
SkillRevise & $+17.71$ & $+10.22$ & $+11.51$ \\
\bottomrule
\end{tabular*}
\caption{Paired reward gains by task-change category.}
\label{tab:variant_gains}
\end{table}

\FloatBarrier
\section{Operational Knowledge and Revision Details}
\label{app:implementation_details}

This appendix specifies script preparation, the knowledge representation, feedback construction, and revision interface
described in Section~\ref{sec:method}.

\paragraph{Preparing executable operations.}
Script construction starts from public task requests and available helper
implementations. A task-family review identifies reusable operations, their
input and output interfaces, required dependencies, and decisions that remain
with the agent. Suitable helpers are reused; missing operations are implemented
as parameterized functions and command-line interfaces. Generated packages
separate implementation files from synthetic unit tests and include usage
examples and declared limitations. Instance-specific answers, fixed task paths,
and hidden-test contents are excluded from the generation input and output.

The generation prompt requests tests covering at least two distinct synthetic
inputs and an invalid input. The component-test runner executes these tests in
a network-disabled container with the package mounted read-only. These checks
exercise helper interfaces; downstream task evaluation remains a separate
stage. Once the script set $P$ is prepared, an authoring call composes $M_0$
from the public request and the actual interfaces in $P$. Subsequent guidance
revision preserves $P$. Appendix~\ref{app:prompts} gives both prompt templates.

\paragraph{Source-grounded knowledge records.}
The extraction input retains the full text of the supplied public scripts, with
numbered source lines. For parsable Python scripts, an AST index records
definitions, parameters, call expressions, conditional branches, and their source
spans; other sources retain their original text. The model produces up to 12
operational knowledge records following Equation~\ref{eq:operational-knowledge}.
Each source reference specifies
a file and inclusive line range. The system checks that each range exists in the
supplied source and attaches the exact code at that location, together with the
source hash.

\paragraph{Workflow comparisons and feedback.}
Given $\mathcal{K}$, the task request, current guidance, and historical trace,
a single model call produces workflow comparisons and associated recommendations.
Each recommendation is associated with one or more comparisons proposing an action;
their knowledge references jointly cover those cited by the recommendation.
The output allows at most 12 comparison rows and 12 feedback items. Feedback
cites exact task-request quotes and, for conflicting instructions,
current-guidance quotes. Validation
checks the knowledge references, comparison--feedback associations, and quoted
text. Appendix~\ref{app:prompts} specifies the output fields.

\paragraph{Targeted edits and candidate formation.}
The revision model returns either no change or up to four non-overlapping edits
across the designated Markdown files. Each edit specifies its file, an exact
original passage, replacement text of at most 8,000 characters, and one or more
feedback identifiers. The system checks editable paths, unique matches for the
original passages, valid feedback references, and non-overlapping spans. Edits
retain some original nonblank text in each modified file; scripts and other files
remain unchanged. Valid edits are applied to form the candidate. Empty feedback,
a no-change response, or invalid revision output retains the current guidance.
The candidate is frozen before downstream execution and task-verifier evaluation;
verifier outcomes are not used to select its edits.

\begin{table*}[!t]
\centering\small
\setlength{\tabcolsep}{5pt}
\begin{tabular}{@{}p{0.15\textwidth}p{0.22\textwidth}p{0.36\textwidth}p{0.20\textwidth}@{}}
\toprule
Condition & Input skill & Update & Downstream evaluation \\
\midrule
Skill-Creator & Generated $S_0=(P,M_0)$ & No subsequent evolution & Three executions of $S_0$ \\
\addlinespace[3pt]
No-verifier & Initial $S_0$ & Method-specific update using the public request, scripts, and unlabeled trace & Freeze output; execute three times \\
\addlinespace[3pt]
Binary / Report / Diagnostic & Initial $S_0$ & Method-specific update with the specified task-verifier feedback; retain $S_0$ on parent PASS & Freeze output; execute three times \\
\addlinespace[3pt]
VISE & Actual incoming evolved $S=(P,M)$ & Extract knowledge from $P$, align it with $M$, and revise $M$ using tool-grounded feedback & Freeze $S'=(P,M')$; execute three times \\
\addlinespace[3pt]
Native & Method's own workflow input & Run the original SkillLearn or SkillRevise workflow, with at most one revision & Freeze selected output; execute three times \\
\bottomrule
\end{tabular}
\caption{Starting skills and update paths in the main evaluation.}
\label{tab:workflow_comparison}
\end{table*}

\section{Main Evaluation: Protocol and Coverage}
\label{app:main_protocol}

\paragraph{Tasks and evaluation.}
The evaluation includes 84 SkillsBench and 99 SkillLearnBench tasks.
Candidate units are task-specific or
shared by tasks using the same skill package.
Each task--method--condition combination has three fixed evaluation records,
with candidates frozen before evaluation.

\paragraph{Native workflows.}
SkillLearn and SkillRevise each run their native workflow once, with at most
one revision. The resulting candidate is selected and frozen before three
downstream executions in separate sessions and clean workspaces.
The Native results cover 83 SkillsBench and 99 SkillLearnBench tasks per method.

\paragraph{Verifier feedback.}
Report contains the parent judgment and up to four recorded test
failures. Available observations are preserved, and missing expected or observed
values remain unspecified. Diagnostic contains the parent judgment,
bounded native-result metadata, and up to eight failure records with diagnostic
messages and traces, within an 8,192-token report limit. Both reports exclude
verifier source and hidden-test source.

\paragraph{Revision interfaces and stopping policy.}
The initial guidance is generated from the public request and script implementations
without human-authored guidance. All conditions receive the same public source
projection and unlabeled parent execution trace. Self-Refine uses
one revision call; TraceEvo, SkillLearn, and SkillRevise use an analysis call
followed by a revision call. SkillLearn and
SkillRevise are adapted to one feedback-conditioned update. No method updates
memory across tasks, and SkillRevise's principle bank remains frozen.
Binary, Report, and Diagnostic retain the initial skill on a recorded parent PASS
and revise only on FAIL. No-verifier and \system{} revise without task-verifier
judgments or this verifier-based stopping rule.

\paragraph{Model and execution settings.}
In the controlled feedback comparison, revision calls use \texttt{gpt-5.5}, temperature zero, and a 16,384-token completion
cap per call. Downstream execution also uses \texttt{gpt-5.5}, retaining each
task's registered execution limits across conditions.

\Needspace{5\baselineskip}
\paragraph{Metrics.}
The non-VISE comparison has sixteen method--feedback cells and one shared
Skill-Creator baseline per task, each with three executions.
For a task with $c$ successful executions, P@1 averages $c/3$, P@3 averages
$\mathbf{1}[c\geq1]$, and $\mathrm{Hit}^{3}$ averages $\mathbf{1}[c=3]$.
For the four non-VISE feedback conditions and Skill-Creator, all three metrics
use the same 183 tasks (84 SkillsBench and 99 SkillLearnBench tasks); VISE uses
the complete-score subsets specified below. Overall weights tasks equally rather
than averaging benchmark percentages.

\paragraph{Starting skills and update paths.}
Table~\ref{tab:workflow_comparison} distinguishes initial skill construction
from the input to a later revision. VISE takes the frozen output of the
preceding evolution stage as its input skill $S=(P,M)$.
The historical trace $\tau$ supplies execution context.

\paragraph{Computational cost.}
Table~\ref{tab:revision_cost} compares recorded token usage and cumulative
API time for skill revision. No-verifier, Binary, Report, and Diagnostic
measure the update from $S_0$; VISE measures the additional refinement of the
incoming evolved skill $S=(P,M)$. Each row averages the same 122 tasks,
including cases that retain the input skill. VISE uses 41.62k--44.79k input
tokens and 1.70k--1.95k output tokens per skill, with API time ranging from
57.97 to 82.27 seconds.

\begin{table}[!t]
\centering\small
\setlength{\tabcolsep}{3pt}
\begin{tabular*}{\columnwidth}{@{\extracolsep{\fill}}llrrr@{}}
\toprule
Method & Feedback & Input (k) & Output (k) & Time (s) \\
\midrule
Self-Refine & No-verifier & 20.38 & 1.92 & 36.97 \\
 & Binary & 14.25 & 1.58 & 26.36 \\
 & Report & 15.17 & 1.37 & 32.58 \\
 & Diagnostic & 15.21 & 1.42 & 26.33 \\
 & VISE & 42.33 & 1.85 & 82.27 \\
\midrule
TraceEvo & No-verifier & 41.02 & 2.98 & 61.70 \\
 & Binary & 28.41 & 2.03 & 40.26 \\
 & Report & 30.32 & 2.00 & 42.18 \\
 & Diagnostic & 30.26 & 1.91 & 41.94 \\
 & VISE & 41.62 & 1.70 & 58.55 \\
\midrule
SkillLearn & No-verifier & 37.57 & 2.68 & 60.61 \\
 & Binary & 26.53 & 1.93 & 37.25 \\
 & Report & 25.45 & 1.70 & 37.47 \\
 & Diagnostic & 28.41 & 1.73 & 38.96 \\
 & VISE & 44.79 & 1.95 & 64.85 \\
\midrule
SkillRevise & No-verifier & 41.33 & 2.66 & 36.92 \\
 & Binary & 28.91 & 1.88 & 24.05 \\
 & Report & 29.74 & 2.06 & 28.38 \\
 & Diagnostic & 30.70 & 2.06 & 29.23 \\
 & VISE & 42.79 & 1.83 & 57.97 \\
\bottomrule

\end{tabular*}
\caption{Average revision-stage token usage and cumulative API time on
122 shared tasks with complete cost records.}
\label{tab:revision_cost}
\end{table}

\begin{figure*}[!t]
\centering
\includegraphics[width=\textwidth]{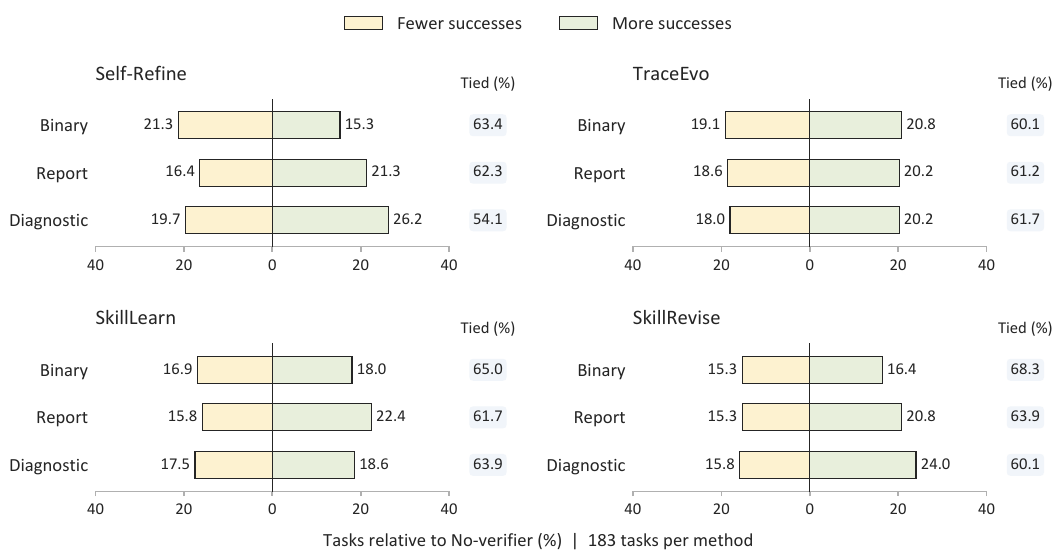}
\caption{Task-level changes relative to No-verifier on 183 tasks per method.
Bars show percentages with fewer (left) or more (right) successes across three
evaluations; side columns show ties.}
\label{fig:verifier_task_effects}
\end{figure*}

\paragraph{Component-comparison protocol.}
\label{app:ablation_protocol}
Table~\ref{tab:component_ablation} uses the 60-task TraceEvo subset
(30 tasks per benchmark), with three records per task and condition.
w/o Workflow Revision uses the corresponding No-verifier records from
Table~\ref{tab:main} on these 60 tasks, without the subsequent VISE revision.
w/o Structured Feedback performs one additional revision using source code and
the unlabeled trace, without structured knowledge or workflow alignment.
Full VISE instead organizes implementation details into operational knowledge
and constructs feedback specifying how the existing guidance should be revised.
The w/o Workflow Alignment condition reuses the
extracted knowledge but removes action-binding records and finding bindings,
retaining the common evidence-source contract and revision validator.

\paragraph{Component success-count distributions.}
Table~\ref{tab:component_success_counts} expands the scores in
Table~\ref{tab:component_ablation} into task counts. Let $n_j$ denote the number
of tasks with $j$ successes among the three evaluation records. Among the
$60-n_0$ tasks solved at least once, the mean success count is
$(n_1+2n_2+3n_3)/(60-n_0)$.

\begin{table}[!htbp]
\centering\small
\setlength{\tabcolsep}{3pt}
\begin{tabular*}{\linewidth}{@{\extracolsep{\fill}}lrrrrr@{}}
\toprule
Condition & $n_0$ & $n_1$ & $n_2$ & $n_3$ & Mean \\
\midrule
\shortstack[l]{w/o Workflow\\Revision} & 14 & 18 & 13 & 15 & 1.93 \\
\addlinespace[3pt]
\shortstack[l]{w/o Structured\\Feedback} & 24 & 8 & 13 & 15 & 2.19 \\
\addlinespace[3pt]
\shortstack[l]{w/o Workflow\\Alignment} & 17 & 18 & 8 & 17 & 1.98 \\
\midrule
Raw-Source Revision & 22 & 14 & 16 & 8 & 1.84 \\
\midrule
Full VISE & 14 & 6 & 22 & 18 & 2.26 \\
\bottomrule
\end{tabular*}
\caption{Success-count distributions for the 60-task TraceEvo component
comparisons. Mean is the number of successes among tasks with at least one
success.}
\label{tab:component_success_counts}
\end{table}

\Needspace{6\baselineskip}
\paragraph{Evaluation coverage.}
For each method, all three VISE metrics use the same set of tasks with complete
evaluation records for both the starting skill and its VISE revision.
Table~\ref{tab:vise_coverage} reports the
number of tasks used for each benchmark.

\begin{table}[!htbp]
\centering\small
\setlength{\tabcolsep}{4pt}
\begin{tabular}{lrrr}
\toprule
Method & SkillsBench & SkillLearnBench & Overall \\
\midrule
Self-Refine & 82/84 & 97/99 & 179/183 \\
TraceEvo & 82/84 & 97/99 & 179/183 \\
SkillLearn & 81/84 & 96/99 & 177/183 \\
SkillRevise & 82/84 & 97/99 & 179/183 \\
\bottomrule

\end{tabular}
\caption{VISE evaluation coverage: tasks with three native scores for both the
starting skill and its revision, relative to the full task set.}
\label{tab:vise_coverage}
\end{table}

\section{Task-Level Effects of Verifier Feedback}
\label{app:verifier_task_effects}
Figure~\ref{fig:verifier_task_effects} decomposes the comparisons into tasks with
more, equal, or fewer successful executions than No-verifier. Each comparison
uses the same 183 tasks and three evaluations per condition.

\paragraph{Category-level comparisons.}
Figure~\ref{fig:verifier_category_effects} uses SkillsBench's official primary
task categories~\citep{li2026skillsbench} and SkillLearnBench's official
task-family categories~\citep{zhong2026skilllearnbench}.
Within each method, all five conditions use the same complete-task set as
Figure~\ref{fig:repeated_success}: 179, 179, 177, and 179 tasks for Self-Refine,
TraceEvo, SkillLearn, and SkillRevise, respectively.
Each cell subtracts the Table~\ref{tab:main} No-verifier P@1 from the feedback
condition's P@1 on the same category tasks, including the VISE comparisons.

\begin{figure*}[!t]
\centering
\includegraphics[width=0.97\textwidth]{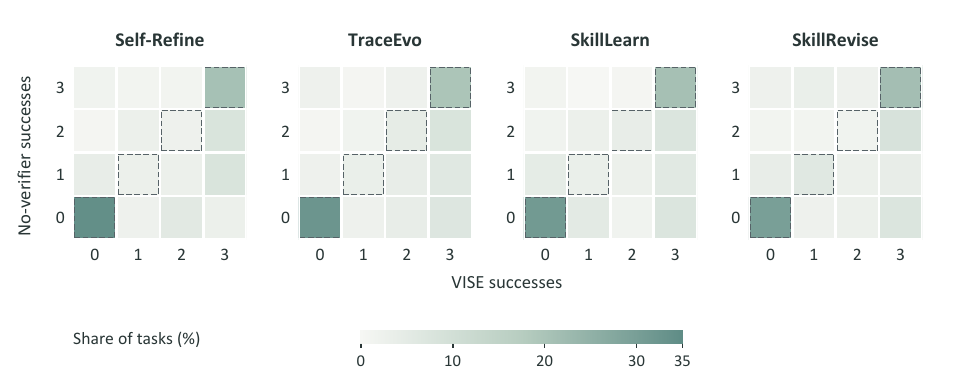}
\caption{Task-level success-count transitions from No-verifier to VISE.
Color encodes the percentage of tasks in each pair of counts, with a shared
scale across methods. Dashed diagonal cells indicate unchanged counts;
cells to their right indicate more successes under VISE, and those to their
left indicate fewer. Cohorts contain 179, 179, 177, and 179 tasks, respectively.}
\label{fig:success_transitions}
\end{figure*}

\begin{figure*}[!t]
\centering
\includegraphics[width=0.97\textwidth]{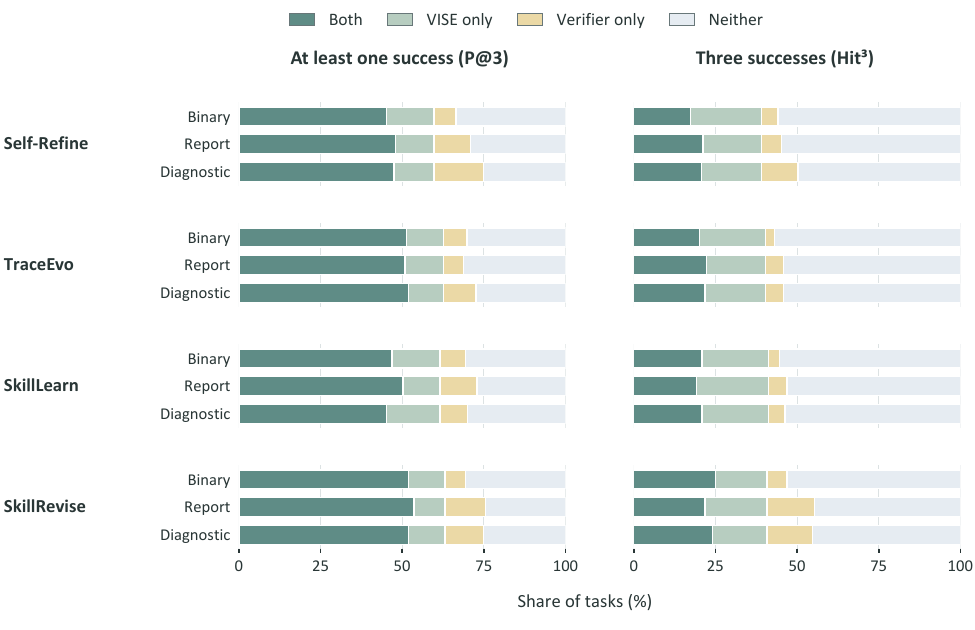}
\caption{Overlap in successful tasks between VISE and task-verifier feedback.
Each bar compares VISE with its labeled feedback condition on the same tasks.
The four segments partition tasks by whether both, only VISE, only the verifier
condition, or neither meets the panel's success criterion. Left: at least one
success; right: three successes. Cohorts match Figure~\ref{fig:success_transitions}.}
\label{fig:success_overlap}
\end{figure*}

\paragraph{Success-count transitions.}
Figure~\ref{fig:success_transitions} pairs each task's No-verifier and VISE
success counts. Of the 33, 37, 36, and 35 tasks newly reaching three successes
under VISE, respectively, 28, 25, 23, and 22 had one or two successes under
No-verifier. Conversely, 7, 9, 3, and 14 tasks move from three successes to
fewer successes. Thus, most newly observed three-success tasks come from the
intermittently successful group rather than the zero-success group.
The method order is Self-Refine, TraceEvo, SkillLearn, and SkillRevise;
both figures below use the same within-method cohorts as the category analysis.

\paragraph{Overlap in successful tasks.}
Figure~\ref{fig:success_overlap} compares the identities of tasks solved by
VISE and each task-verifier feedback condition. Under the three-success
criterion, VISE-only tasks outnumber verifier-only tasks in all twelve
comparisons. The ordering can differ for at-least-once success: for
Self-Refine versus Diagnostic, 22 tasks succeed only under VISE and 27 only
under Diagnostic, whereas the corresponding three-success counts are 33 and
20. Task coverage and repeated success therefore distinguish different
sets of tasks within the same comparison.

\FloatBarrier
\section{Prompt Templates}
\label{app:prompts}

The following boxes present compact forms of the implemented prompt interfaces.
Task requests, source files, guidance, and traces appear as input placeholders;
output field names follow the implementation. The templates share a data rule:
embedded source, requests, guidance, and observations are treated as data rather
than instructions, and task outcomes or hidden answers must not be invented.

\begin{nolinenumbers}
\begingroup
\tcbset{visepromptlayout/.style={breakable=false}}

\begin{visepromptbox}[viseauthorframe]{Constructing Missing Scripts}
\textbf{Instruction.} Implement reusable helpers for the supplied public
workflow. Expose explicit inputs and outputs through parameterized functions
and command-line interfaces. Preserve original inputs and report unsupported
cases. Keep task-specific values configurable; do not embed answers, hidden
tests, fixed task paths, or task assets in the package. Generate implementations
and synthetic tests separately from \skillmd{}.

\textbf{Inputs.} Public task-family requests and the operation review,
including proposed interfaces, dependencies, and limitations.

\textbf{Output.} Return a JSON package containing \texttt{name},
\texttt{description}, \texttt{dependencies}, \texttt{files},
\texttt{usage\_examples}, and \texttt{limitations}. Each file has a path and
content. Place implementations under \texttt{scripts/} and synthetic unit
tests under \texttt{tests/}. Cover at least two distinct inputs and an invalid
input. Document CLI usage. Do not perform network access, invoke shell commands,
read credentials, install packages, or wrap an LLM invocation in the helper.
\end{visepromptbox}

\begin{visepromptbox}[viseauthorframe]{Initial Workflow Guidance}
\textbf{Instruction.} Author the initial reusable skill documentation for an
executable skill package. Use the task request as an example of intended use.
Describe general procedures and the real interfaces exposed by the read-only
scripts; do not solve the task or embed its answer. Write concise guidance that
helps a future agent decide when and how to use the scripts. Refer to paths,
symbols, arguments, outputs, and workflows supported by the supplied source.

\textbf{Inputs.} \texttt{[CREATOR\_SPECIFICATION]},
\texttt{[AUTHORING\_ADAPTER]}, \texttt{[TASK\_REQUEST]},
\texttt{[AUTHORIZED\_TARGETS]}, and \texttt{[PUBLIC\_SCRIPTS]}.
Each target specifies a Markdown path and required skill name.

\textbf{Output.} Call \texttt{submit\_initial\_skill\_package} with
\texttt{summary} and \texttt{documents}. Each document contains
\texttt{path} and \texttt{content}. Return every authorized document exactly
once, preserving its path and required name. Begin each document with YAML
front matter containing \texttt{name} and \texttt{description}.
Scripts remain unchanged.
\end{visepromptbox}

\begin{visepromptbox}[viseknowledgeframe]{Operational Knowledge Extraction}
\textbf{Instruction.} Extract reusable operational knowledge from the supplied
scripts before diagnosing a task. Describe capabilities, when to use them,
ordered steps, and conditional limitations. Read implementations in context,
not just names or syntax nodes. Group related operations as workflows rather
than producing one record per node. An available helper need not be mandatory,
and a default value need not be wrong.

\textbf{Input.} \texttt{[SOURCE\_PACKET]} contains public script paths,
numbered source lines, and syntax indices where parsing is supported. Other
sources retain their text.

\textbf{Output.} Return a JSON object with a \texttt{workflows} array of at
most 12 records, each containing \texttt{id}, \texttt{capability},
\texttt{applicability}, \texttt{steps}, \texttt{constraints}, and
\texttt{sources}. Each source reference contains \texttt{path},
\texttt{start\_line}, and \texttt{end\_line}. Select narrow, inclusive ranges
using the supplied line numbers; the host attaches their exact source text.
An empty array is allowed. Infer applicability and usage advice from the
cited implementation.
\end{visepromptbox}

\begin{visepromptbox}[visefeedbackframe]{Workflow-Grounded Feedback}
\textbf{Instruction.} Align the public workflow with the supplied guidance
across all authorized Markdown files. Organize comparisons by workflows.
Distinguish an existing operation, a known limitation, and an unfinished
transition when that limitation applies. Equivalent operations across files
are covered; stating a limitation alone does not supply an alternative path.
Recommend a necessary remaining operation under its applicable conditions,
rather than a restatement or stylistic change. Preserve useful guidance.

\textbf{Inputs.} \texttt{[TASK\_REQUEST]}, \texttt{[SKILL\_MDS]},
\texttt{[OPERATIONAL\_KNOWLEDGE]}, and \texttt{[HISTORICAL\_TRACE]}.
\texttt{SKILL\_MDS} maps authorized paths to their current text.

\textbf{Output.} Return a JSON object with \texttt{comparisons}.
Each comparison contains \texttt{alignment} rows and either one
\texttt{feedback} item or \texttt{null}. Use 1--12 alignment rows in total
and at most 12 feedback items.

\textbf{Alignment fields.} \texttt{requirement},
\texttt{existing\_operation}, \texttt{existing\_paths},
\texttt{known\_limitation}, \texttt{knowledge\_ids},
\texttt{remaining\_transition}, and \texttt{decision}.
Choose \texttt{COVERED}, \texttt{ADD\_OPERATION}, \texttt{ADD\_FALLBACK},
\texttt{CORRECT\_OPERATION}, or \texttt{UNKNOWN}.
\texttt{COVERED} requires an existing operation and an authorized Markdown
path; use \texttt{UNKNOWN} when that basis is absent.

\textbf{Feedback fields.} \texttt{id}, \texttt{kind},
\texttt{knowledge\_ids}, \texttt{request\_quotes},
\texttt{skill\_quotes}, \texttt{event\_ids}, \texttt{issue},
\texttt{recommendation}, and \texttt{applicability}.
Use \texttt{MISSING\_GUIDANCE} or \texttt{CONFLICTING\_GUIDANCE}.
Each item cites supplied knowledge and an exact request quotation;
conflicts additionally cite exact passages in the supplied Markdown.
Set \texttt{event\_ids} to an empty array.

\textbf{Association rule.} Each feedback item is supported by at least one
action row with a remaining transition and overlapping knowledge identifiers.
Together, its action rows cover all knowledge identifiers cited by the item.
\texttt{COVERED} and \texttt{UNKNOWN} rows may provide context but do not
support a proposed change. Return \texttt{null} when the comparison does not
support a useful revision.
\end{visepromptbox}

\tcbset{visepromptlayout/.style={breakable=false,fontupper=\small,top=6pt,bottom=6pt}}
\begin{visepromptbox}[viserevisionframe]{Feedback-Guided Workflow Revision}
\textbf{Instruction.} Improve only the authorized Markdown using the supplied
conditional feedback. Evaluate the recommendations and preserve useful
existing instructions. Do not turn optional capabilities into unconditional
requirements or paste all extracted knowledge into the skill. A grounded
guidance gap can justify an edit without a runtime exception. Scripts and
all other files are read-only.

\textbf{Inputs.} \texttt{[TASK\_REQUEST]}, \texttt{[SKILL\_MDS]},
\texttt{[OPERATIONAL\_KNOWLEDGE]}, \texttt{[FEEDBACK]}, and
\texttt{[HISTORICAL\_TRACE]}.

\textbf{Output.} Return \texttt{decision: KEEP} if no change is useful,
or \texttt{decision: EDIT} with an \texttt{edits} array. Each edit contains
\texttt{path}, \texttt{old}, \texttt{new}, and \texttt{finding\_ids}.
The original passage must be a unique exact substring in its authorized file;
\texttt{finding\_ids} references the feedback addressed by the edit.
Return at most four non-overlapping edits across all authorized files, retaining
some original nonblank text in each edited file.
\end{visepromptbox}

\begin{visepromptbox}[viseablationframe]{w/o Structured Feedback}
\textbf{Instruction.} Revise the skill guidance for future use based on the
task, public source scripts, and historical execution trace. Identify concrete
improvements yourself. Scripts and assets are read-only. Preserve names and
front matter, and do not replace an entire guidance file.

\textbf{Inputs.} \texttt{[TASK\_REQUEST]}, \texttt{[SKILL\_MDS]},
\texttt{[PUBLIC\_SCRIPTS]}, and \texttt{[HISTORICAL\_TRACE]}, without a
structured feedback package or task-verifier results.

\textbf{Output.} Return \texttt{decision: KEEP}, or
\texttt{decision: EDIT} with at most four non-overlapping edits.
Each edit contains \texttt{path}, \texttt{old}, and \texttt{new}.
Each original span occurs exactly once in its file; each replacement contains
at most 8,000 characters.
\end{visepromptbox}

\newpage
\begin{visepromptbox}[viseablationframe]{w/o Workflow Alignment}
\textbf{Instruction.} Compare script-derived operational knowledge with the
public request and the authorized Markdown. Provide actionable, conditional
feedback rather than a source summary. Identify relevant missing instructions
or instructions that conflict with the implementation. Preserve useful
guidance already present.

\textbf{Inputs.} \texttt{[TASK\_REQUEST]}, \texttt{[SKILL\_MDS]},
\texttt{[OPERATIONAL\_KNOWLEDGE]}, and \texttt{[HISTORICAL\_TRACE]}.

\textbf{Output.} Return a \texttt{feedback} array directly, without
alignment rows or action-to-feedback associations. Use the same feedback
fields, knowledge references, and exact-quotation requirements as the full
feedback interface. Use \texttt{MISSING\_GUIDANCE} or
\texttt{CONFLICTING\_GUIDANCE}, with empty \texttt{event\_ids}.
Return at most 12 items, or an empty array if no useful difference is supported.
The resulting feedback uses the same workflow-revision interface.
\end{visepromptbox}

w/o Workflow Revision uses the No-verifier skill directly and therefore
introduces no additional VISE prompt.
\endgroup
\end{nolinenumbers}

\clearpage
\begingroup
\raggedright
\tcbset{visepromptlayout/.style={breakable=false,fontupper=\small}}
\makeatletter
\renewcommand{\paragraph}{\@startsection{paragraph}{4}{\z@}%
  {.8ex \@plus .2ex}{.25ex}{\normalfont\normalsize\bfseries}}
\makeatother
\section{A Complete Workflow-Revision Example}
\label{app:revision_case}
This appendix expands the video object-counting example in
Section~\ref{sec:rq3_implementation}, following one Self-Refine skill from its
incoming guidance to its revised execution. The task requires counting objects
in video frames and recording the counts in a CSV file. The executable counter
is identical before and after revision; the change is confined to
\skillmd{}.\looseness=-1

\paragraph{Implementation evidence.}
The file \texttt{scripts/count\_objects.py} exposes five required arguments:
\texttt{--tool}, \texttt{--input\_image}, \texttt{--object\_image},
\texttt{--threshold}, and \texttt{--dedup\_min\_dist} (lines 25--44).
Its counting function reads both images in grayscale, applies normalized
template matching, filters matches by the threshold, and removes nearby
duplicates (lines 47--73). The command-line entry point prints the resulting
integer count with the input and template paths (lines 76--86).\looseness=-1

\paragraph{Operational knowledge.}
VISE groups these source spans into the record
\texttt{count-objects-cli}. Its capability is template-based object counting;
its conditions require readable input and template images. The record connects
input preparation, configurable matching conditions, and output retrieval into
a procedure for obtaining a count usable by subsequent workflow steps.\looseness=-1

\paragraph{Guidance gap and feedback.}
The incoming guidance says only to inspect scripts and use runtime evidence.
The workflow comparison identifies the missing counting procedure and emits
\texttt{finding-1}, linked to \texttt{count-objects-cli} and the task's
object-counting requirement. It recommends adding the invocation procedure
and instructions to use the printed count for each frame--template pair.
Frame extraction and CSV writing are separate
operations outside the counter.\looseness=-1

\newpage
\begin{nolinenumbers}
\begin{visepromptbox}[viseauthorframe]{Incoming guidance: exact excerpt}
Inspect the files under \texttt{scripts/} to determine their interfaces and
behavior. Use runtime evidence when available.
\end{visepromptbox}

\begin{visepromptbox}[viserevisionframe]{Added guidance: exact excerpts}
\textbf{Counting objects with a template}

When a task asks to count occurrences of a sample object image in another
image, the package provides a \texttt{count} tool in
\texttt{scripts/count\_objects.py}.

\ldots

Choose threshold and de-duplication distance based on the image/template scale
and validate as needed. The command prints a line reporting the number of
objects for that input image and template; use that reported count for each
frame/template pair.
\end{visepromptbox}
\end{nolinenumbers}

\paragraph{Accepted revision.}
One edit appends the counting procedure to \texttt{object\_counter/SKILL.md}
and cites \texttt{finding-1}, preserving the existing instructions.
Numerical thresholds and frame selection remain execution-time choices.\looseness=-1

\paragraph{Use in execution.}
Both recorded executions invoke the counter, with different frame selections
and matching parameters. In the incoming-skill execution,
four calls inspect a single frame after the CSV file has already been written.
In the revised-skill execution, calls occur inside a frame--template loop:
\texttt{subprocess.check\_output} captures the printed result, a regular
expression extracts its integer count, and counts are aggregated into
\texttt{coins} and \texttt{turtles} before \texttt{csv.DictWriter} writes five
rows with \texttt{frame\_id}. The revised execution uses the counter's outputs
to construct the final CSV.\looseness=-1
\par
\endgroup

\onecolumn
\section{Main-Result Task Heatmaps}
\label{app:main_task_heatmaps}

\begin{figure}[!ht]
\centering
\includegraphics[width=\textwidth]{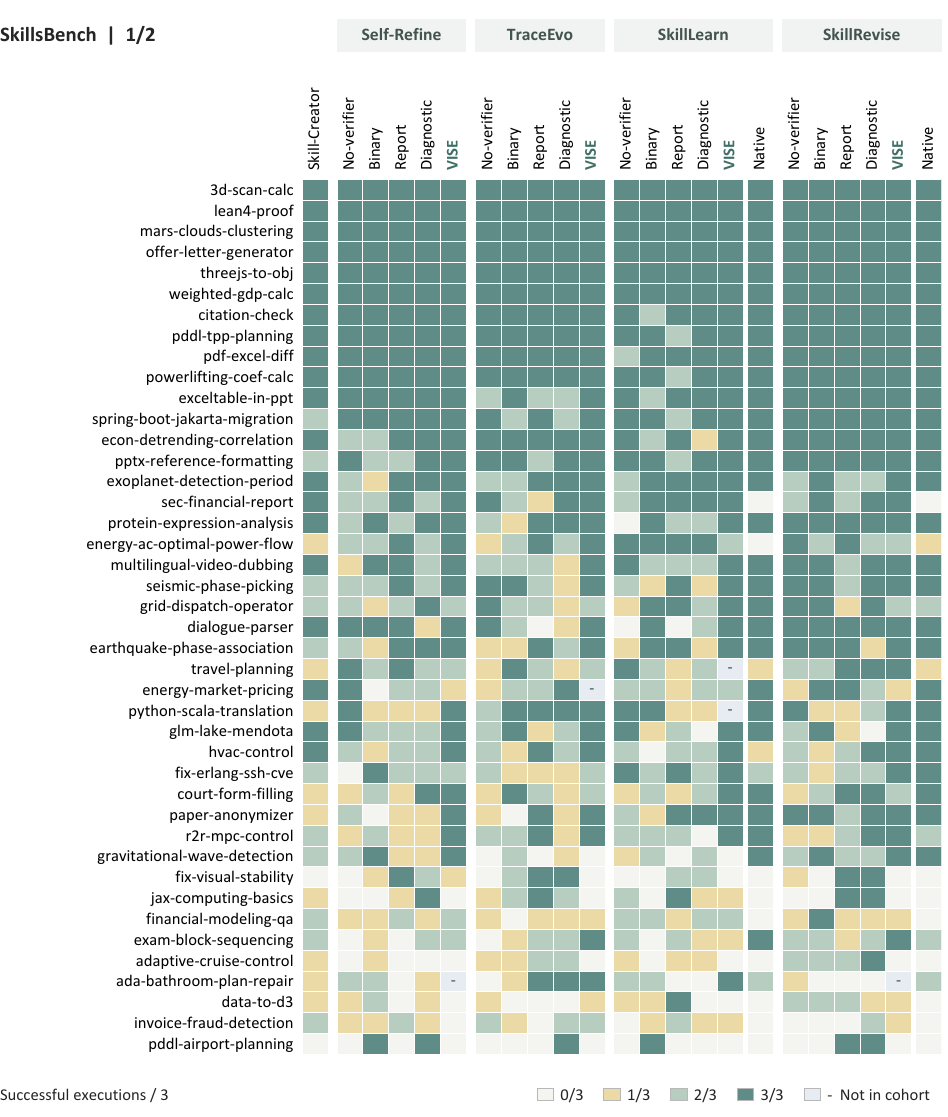}
\caption{Task-level outcomes on SkillsBench (part 1 of 2). Each cell shows successful executions out of three for a condition in Table~\ref{tab:main}; a dash marks a task outside that condition's evaluation cohort. Tasks are sorted by total successes across Skill-Creator and the sixteen non-VISE controlled conditions, with the same order across columns. Native denotes each method's original workflow.}
\label{fig:main_tasks_skillsbench_1}
\end{figure}
\clearpage

\begin{figure}[!ht]
\centering
\includegraphics[width=\textwidth]{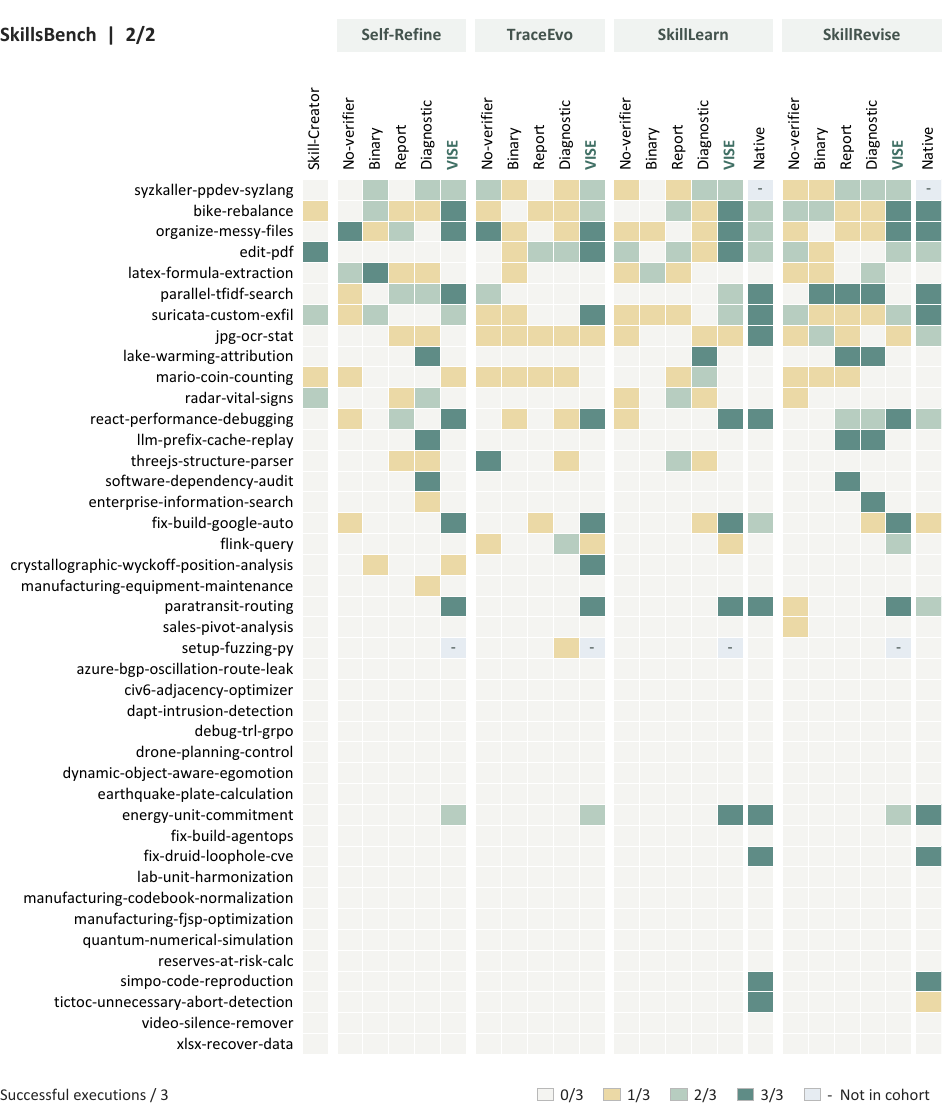}
\caption{Task-level outcomes on SkillsBench (part 2 of 2). Cells show successful executions out of three; dashes mark tasks outside the corresponding evaluation cohort. Columns and task ordering follow Figure~\ref{fig:main_tasks_skillsbench_1}.}
\label{fig:main_tasks_skillsbench_2}
\end{figure}
\clearpage

\begin{figure}[!ht]
\centering
\includegraphics[width=\textwidth]{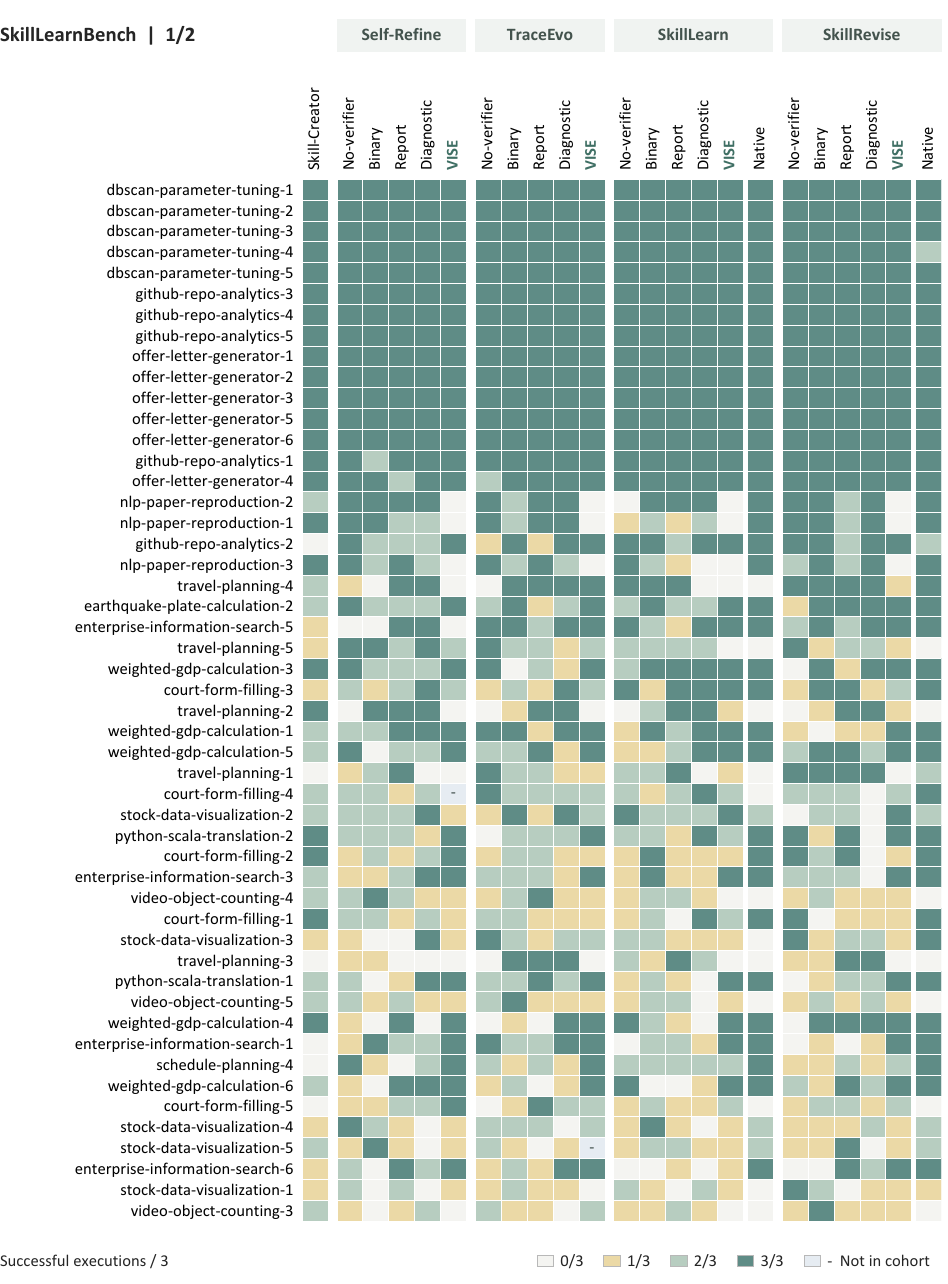}
\caption{Task-level outcomes on SkillLearnBench (part 1 of 2). Cells show successful executions out of three; dashes mark tasks outside the corresponding evaluation cohort. Tasks are sorted by total successes across Skill-Creator and the sixteen non-VISE controlled conditions, using the same columns as Figure~\ref{fig:main_tasks_skillsbench_1}.}
\label{fig:main_tasks_skilllearnbench_1}
\end{figure}
\clearpage

\begin{figure}[!ht]
\centering
\includegraphics[width=\textwidth]{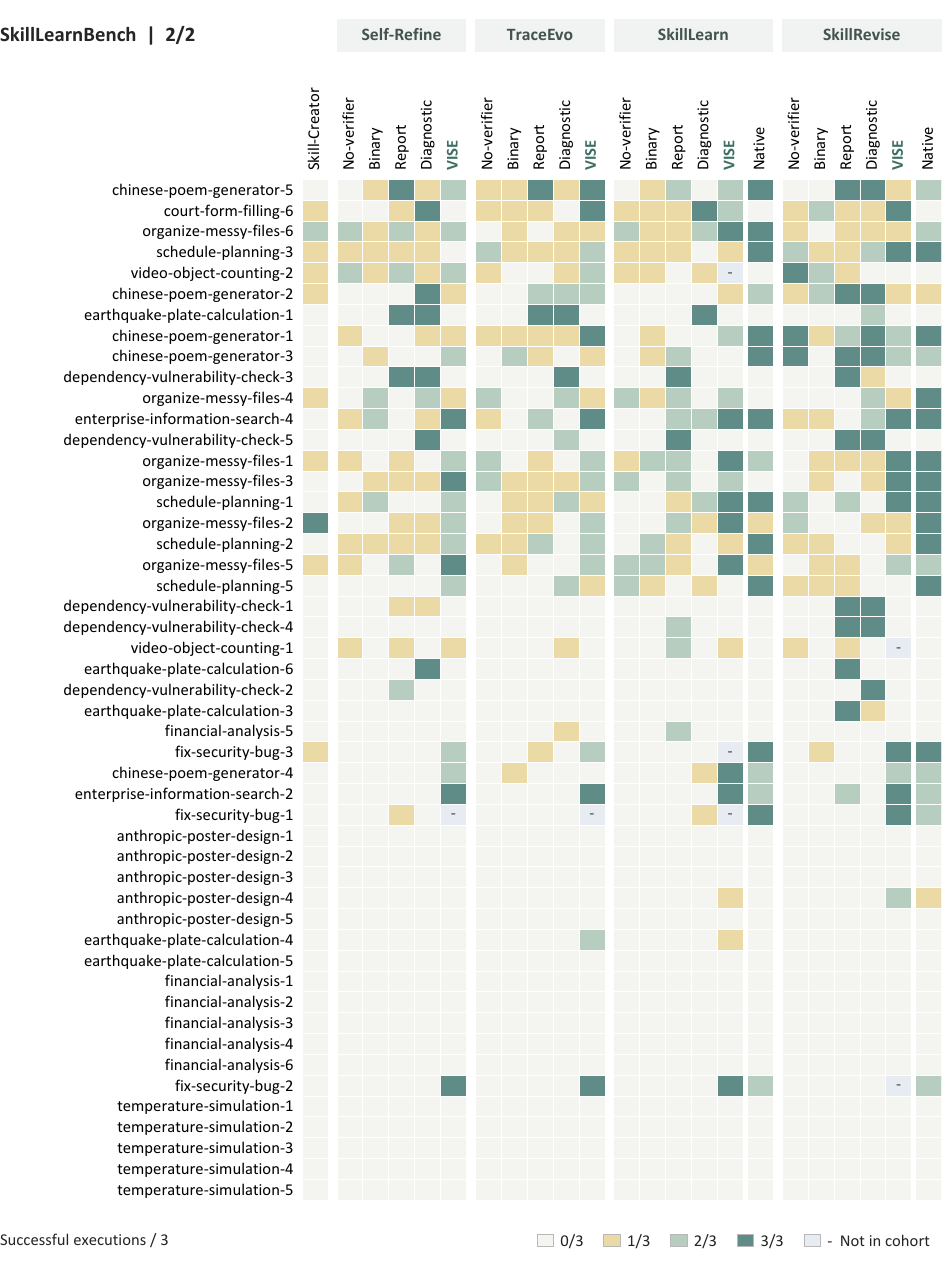}
\caption{Task-level outcomes on SkillLearnBench (part 2 of 2). Cells show successful executions out of three; dashes mark tasks outside the corresponding evaluation cohort. Columns and task ordering follow Figure~\ref{fig:main_tasks_skilllearnbench_1}.}
\label{fig:main_tasks_skilllearnbench_2}
\end{figure}

\end{document}